\documentclass[twocolumn]{article}

\usepackage{subfigure}
\usepackage{amsmath}
\usepackage{amsfonts}
\usepackage{amssymb}
\usepackage[ruled,linesnumbered]{algorithm2e}
\usepackage{amsthm}
\usepackage{verbatim}
\usepackage{enumerate}
\usepackage{graphicx}

\usepackage{cite}
\newcommand{\citet}{\cite}
\newcommand{\citep}{\cite}

\newcommand{\bm}[1]{\mathbf{#1}}
\newcommand{\tsize}{m}
\newcommand{\fsize}{n}
\newcommand{\desiredfcount}{k}

\newcommand{\xmatrix}{X}

\newcommand{\kernelm}{K}

\newcommand{\learningf}{f}
\newcommand{\lossfunction}{l}

\newcommand{\dsikm}{G}
\newcommand{\labelvec}{\bm{y}}
\newcommand{\leftvec}{\bm{u}}
\newcommand{\rightvec}{\bm{v}}

\newcommand{\dualvec}{\bm{a}}
\newcommand{\primalvec}{\bm{w}}

\newcommand{\regparam}{\lambda}

\newcommand{\idmatrix}{I}

\newcommand{\transpose}{^\textrm{T}}

\newcommand{\anymatrix}{M}
\newcommand{\cachematrix}{C}

\newcommand{\diagvec}{\bm{d}}

\newcommand{\selectedfeatures}{\mathcal{S}}
\newcommand{\candidatefeaset}{\mathcal{R}}
\newcommand{\looindset}{\mathcal{L}}
\newcommand{\looerr}{e_i}
\newcommand{\bestloo}{e}
\newcommand{\bestfeaind}{b}
\newcommand{\pred}{p}

\newcommand{\trainalg}{t}

\newcommand{\qvector}{\bm{q}}

\DeclareMathOperator*{\argmin}{argmin}

\begin{document}

\title{Linear Time Feature Selection for Regularized Least-Squares}

\author{Tapio~Pahikkala, Antti~Airola, and Tapio~Salakoski}

\maketitle

\begin{abstract}
We propose a novel algorithm for greedy forward feature selection for regularized least-squares (RLS) regression and classification, also known as the least-squares support vector machine or ridge regression. The algorithm, which we call greedy RLS, starts from the empty feature set, and on each iteration adds the feature whose addition provides the best leave-one-out cross-validation performance. Our method is considerably faster than the previously proposed ones, since its time complexity is linear in the number of training examples, the number of features in the original data set, and the desired size of the set of selected features. Therefore, as a side effect we obtain a new training algorithm for learning sparse linear RLS predictors which can be used for large scale learning. This speed is possible due to matrix calculus based short-cuts for leave-one-out and feature addition. We experimentally demonstrate the scalability of our algorithm and its ability to find good quality feature sets.
\end{abstract}

\section{Introduction}

Feature selection is one of the fundamental tasks in machine learning.
Simply put, given a set of features, the task is to select an informative subset.
The selected set can be informative in the sense that it guarantees a good
performance of the machine learning method or that it will provide new
knowledge about the task in question. Such increased understanding of the data is sought
after especially in life sciences, where feature selection is often used, for example,
to find genes relevant to the problem under consideration. Other benefits
include compressed representations of learned predictors and faster prediction
time. This can enable the application of learned models when constrained
by limited memory and real-time response demands, as is typically the case
in embedded computing, for example.

The feature selection methods are divided by \citet{guyon2003introduction} into three
classes, the so called filter, wrapper, and embedded methods. In the filter
model feature selection is done as a preprocessing step by measuring only the
intrinsic properties of the data. In the wrapper model \citep{kohavi1997wrappers} the features are
selected through interaction with a machine learning method. The quality of
selected features is evaluated by measuring some estimate of the generalization
performance
of a prediction model constructed using them. The embedded methods minimize an objective function
which is a trade-off between a goodness-of-fit term and a penalty term for a large
number of features (see e.g. \citet{keerthi2007fasttrack}).

On one hand, the method we propose in this work is a wrapper method, since
equivalent results can be obtained by using a black-box learning algorithm
together with the considered search and performance estimation strategies.
On the other hand it may also be considered as an embedded method, because the
feature subset selection is so deeply integrated in the learning process, that
our work results in a novel efficient training algorithm for the method.

As suggested by \cite{john94irrelevant}, we estimate the generalization
performance of models learned with different subsets of features with
cross-validation. More specifically, we consider the leave-one-out
cross-validation (LOO) approach, where each example in turn is left
out of the training set and used for testing. Since the number of possible
feature combinations grows exponentially with the number of features, a search
strategy over the power set of features is needed. The strategy we choose is the greedy
forward selection approach. The method starts from the empty feature set, and
on each iteration adds the feature whose addition provides the best
leave-one-out cross-validation performance. That is, it performs a steepest
descent hill-climbing in the space of feature subsets of size up to a given
limit $\desiredfcount\in\mathbb{N}$ and is greedy in the sense that it does
not remove features once they have been selected. 

Our method is built upon the Regularized least-squares (RLS), also known as the
least-squares support vector machine (LS-SVM) and ridge regression, which is  is a state-of-the art machine learning method suitable both for regression and classification
\cite{hoerl1970ridge,poggio1990networks,saunders1998rrdual,suykens1999lssvm,
suykens2002lssvmbook, rifkin2003rls, poggio2003mathematics}. An important property of the algorithm is that it has a closed form solution,
which can be fully expressed in terms of matrix operations. This allows
developing efficient computational shortcuts for the method, since small
changes in the training data matrix correspond to low-rank changes in the
learned predictor. Especially it makes possible the development of efficient
cross-validation algorithms. An updated predictor, corresponding
to a one learned from a training set from which one example has
been removed, can be obtained via a well-known computationally efficient
short-cut (see e.g. \cite{vapnik1979estimation,wahba1990spline,green1994nonparametric})
which in turn enables the fast computation of LOO-based performance estimates.
Analogously to removing the effect of training examples from learned RLS
predictors, the effects of a feature can be added or removed by updating the
learned RLS predictors via similar computational short-cuts.

Learning a linear RLS predictor with $\desiredfcount$ features and $\tsize$
training examples requires
$O(\min\{\desiredfcount^2\tsize,\desiredfcount\tsize^2\})$ time, since the
training can be performed either in primal or dual form depending whether
$\tsize>\desiredfcount$ or vice versa.
Given that the computation of LOO performance requires $\tsize$ retrainings,
that the forward selection goes through of the order of $O(\fsize)$ features
available for selection in each iteration, and that the forward selection has $\desiredfcount$
iterations, the overall time complexity of the forward selection with LOO
criterion becomes
$O(\min\{\desiredfcount^3\tsize^2\fsize,\desiredfcount^2\tsize^3\fsize\})$
in case RLS is used as a black-box method.

In machine learning and statistics literature, there have been several studies
in which the computational short-cuts for LOO, as well as other types of
cross-validation, have been used to speed up the evaluation of 
feature subset quality for ordinary non-regularized least-squares (see e.g
\citet{shao1993linearcv,zhang1993mcv}) and for RLS
(see e.g. \citet{tang2006gene,ying2006fastloo}). However, the considered approaches
are still be computationally quite demanding, since the LOO
estimate needs to be re-calculated from scratch for each considered subset of
features. Recently, \citet{ojeda2008lssvmselection} introduced a method which
uses additional computational short-cuts for efficient updating of the LOO
predictions when adding new features to those already selected.
The computational cost of the incremental forward
selection procedure is only $O(\desiredfcount\tsize^2\fsize)$ for the method.
The speed improvement is notable especially in cases, where the aim is to select a large number of features but
the training set is small. However, the method is still impractical for
large training sets due to its quadratic scaling.

In this paper, we improve upon the above results by showing how
greedy forward selection according to the LOO criterion can
be carried out in $O(\desiredfcount\tsize\fsize)$ time. Thus, our algorithm,
which we call greedy RLS or greedy LS-SVM, is linear in the number of examples, features, and selected
features, while it still provides the same solution as the straightforward
wrapper approach. As a side effect we obtain a novel training algorithm for
linear RLS predictors that is suitable for large-scale learning and guarantees
sparse solutions. Computational experiments demonstrate the scalability
and efficiency of greedy RLS, and the predictive power of selected
feature sets.

\section{Regularized Least-Squares}\label{RegFramework}

We start by introducing some notation. Let $\mathbb{R}^{\tsize}$ and $\mathbb{R}^{\fsize\times\tsize}$, where $\fsize,\tsize\in\mathbb{N}$, denote the sets of real valued column vectors and $\fsize\times\tsize$-matrices, respectively. To denote real valued matrices and vectors we use capital letters and bold lower case letters, respectively. Moreover, index sets are denoted with calligraphic capital letters. By denoting $\anymatrix_i$, $\anymatrix_{:,j}$, and $\anymatrix_{i,j}$, we refer to the $i$th row, $j$th column, and $i,j$th entry of the matrix $\anymatrix\in\mathbb{R}^{\fsize\times\tsize}$, respectively. Similarly, for index sets $\candidatefeaset\in\{1,\ldots,\fsize\}$ and $\looindset\in\{1,\ldots,\tsize\}$, we denote the submatrices of $\anymatrix$ having their rows indexed by $\candidatefeaset$, the columns by $\looindset$, and the rows by $\candidatefeaset$ and columns by $\looindset$ as $\anymatrix_{\candidatefeaset}$, $\anymatrix_{:,\looindset}$, and $\anymatrix_{\candidatefeaset,\looindset}$, respectively. We use an analogous notation also for column vectors, that is, $\bm{v}_{i}$ refers to the $i$th entry of the vector $\bm{v}$.

Let $\xmatrix\in\mathbb{R}^{\fsize\times\tsize}$ be a matrix containing the
whole feature representation of the examples in the training set, where $\fsize$ is the total number of features and $\tsize$ is the number of training examples. The $i,j$th entry of $\xmatrix$ contains the value of the $i$th feature in the $j$th training example. Moreover, let $\labelvec\in\mathbb{R}^{\tsize}$ be a vector containing the labels of the training examples. In binary classification, the labels can be restricted to be either $-1$ or $1$, for example, while they can be any real numbers in regression tasks.

In this paper, we consider linear predictors of type
\begin{equation}\label{primalsol}
\learningf(\bm{x})=\primalvec\transpose\bm{x}_\selectedfeatures,
\end{equation}
where $\primalvec$ is the $\arrowvert\selectedfeatures\arrowvert$-dimensional vector representation of the learned predictor and $\bm{x}_\selectedfeatures$ can be considered as a mapping of the data point $x$ into $\arrowvert\selectedfeatures\arrowvert$-dimensional feature space.\footnote{In the literature, the formula of the linear predictors often also contain a bias term. Here, we assume that if such a bias is used, it will be realized by using an extra constant valued feature in the data points.}
Note that the vector $\primalvec$ only contains entries corresponding to the features indexed by $\selectedfeatures$. The rest of the features of the data points are not used in prediction phase. The computational complexity of making predictions with (\ref{primalsol}) and the space complexity of the predictor are both $O(\desiredfcount)$, where $\desiredfcount$ is the desired number of features to be selected, provided that the feature vector representation $\bm{x}_\selectedfeatures$ for the data point $x$ is given.

Given training data and a set of feature indices $\selectedfeatures$, we find $\primalvec$ by minimizing
the so-called regularized least-squares (RLS) risk. Minimizing the RLS risk can be expressed as the following problem:
\begin{equation}\label{primalproblem}
\argmin_{\primalvec\in\mathbb{R}^{\arrowvert\selectedfeatures\arrowvert}}
((\primalvec\transpose\xmatrix_\selectedfeatures)\transpose-\labelvec)\transpose((\primalvec\transpose\xmatrix_\selectedfeatures)\transpose-\labelvec) + \regparam\primalvec\transpose\primalvec.
\end{equation}
The first term in (\ref{primalproblem}), called the empirical risk, measures how well the prediction function fits to the training data. The second term is called the regularizer and it controls the tradeoff between the loss on the training set and the complexity of the prediction function.

A straightforward approach to solve (\ref{primalproblem}) is to set the derivative of the objective function with respect to $\primalvec$ to zero. Then, by solving it with respect to $\primalvec$, we get 
\begin{equation}\label{firstform}
\primalvec=(\xmatrix_\selectedfeatures(\xmatrix_\selectedfeatures)\transpose+\regparam\idmatrix)^{-1}\xmatrix_\selectedfeatures\labelvec,
\end{equation}
where $\idmatrix\in\mathbb{R}^{\tsize\times\tsize}$ is the identity matrix.
We note (see e.g. \cite{searle1982matrixalgebra}) that an equivalent result can be obtained from
\begin{equation}\label{secondform}
\primalvec=\xmatrix_\selectedfeatures((\xmatrix_\selectedfeatures)\transpose\xmatrix_\selectedfeatures+\regparam\idmatrix)^{-1}\labelvec.
\end{equation}
If the size of the set $\selectedfeatures$ of currently selected features is smaller than the number of training examples $\tsize$, it is computationally beneficial to use the form (\ref{firstform}) while using (\ref{secondform}) is faster in the opposite case. Namely, the computational complexity of the former is $O(\arrowvert\selectedfeatures\arrowvert^3+\arrowvert\selectedfeatures\arrowvert^2\tsize)$, while the that of the latter is $O(\tsize^3+\tsize^2\arrowvert\selectedfeatures\arrowvert)$, and hence the complexity of training a predictor is $O(\min\{\arrowvert\selectedfeatures\arrowvert^2\tsize,\tsize^2\arrowvert\selectedfeatures\arrowvert\})$.

To support the considerations in the following sections, we introduce the dual form of the prediction function and some extra notation. According to \citet{saunders1998rrdual}, the prediction function (\ref{primalsol}) can be represented in dual form as follows
\begin{equation*}
\learningf(\bm{x})=\dualvec\transpose(\xmatrix_\selectedfeatures)\transpose\bm{x}_\selectedfeatures.
\end{equation*}
Here $\dualvec\in\mathbb{R}^{\tsize}$ is the vector of so-called dual variables, which can be obtained from
\begin{equation*}
\dualvec=\dsikm\labelvec,
\end{equation*}
where
\begin{equation}\label{gmatrixdef}
\dsikm = (\kernelm+\regparam\idmatrix)^{-1}
\end{equation}
and
\begin{equation}\label{fullkernelmatrix}
\kernelm
=(\xmatrix_{\selectedfeatures})\transpose\xmatrix_{\selectedfeatures}.
\end{equation}
Note that if $\selectedfeatures=\emptyset$, $\kernelm$ is defined to be a matrix whose every entry is equal to zero.

In the context of kernel-based learning algorithms (see e.g. \citet{muller2001anintroduction,scholkopf2002kernels,shawetaylor2004kernel}), $\kernelm$ and $\dualvec$ are usually called the kernel matrix and the vector of dual variables, respectively. The kernel matrix contains the inner products between the feature vectors of all training data points. With kernel methods, the feature vector representation of the data points may be even infinite dimensional as long as there is an efficient, although possibly implicit, way to calculate the value of the inner product and the dual variables are used to represent the prediction function. However, while the dual representation plays an important role in the considerations of this paper, we restrict our considerations to feature representations of type $\bm{x}_\selectedfeatures$ which can be represented explicitly.

Now let us consider some well-known efficient approaches for evaluating the LOO performance of a trained RLS predictor (see e.g. \cite{vapnik1979estimation,wahba1990spline,green1994nonparametric,rifkin2007notes} and for the non-regularized case, see e.g \citet{shao1993linearcv,zhang1993mcv}). The LOO prediction for the $j$th training example can be obtained in constant time from
\begin{equation}\label{fastprimalloo}
(1-\qvector_{j})^{-1}(\bm{f}_{j}-\qvector_{j}\labelvec_{j}),
\end{equation}
where
\begin{equation*}
\bm{f}=(\primalvec\transpose\xmatrix_\selectedfeatures)\transpose
\end{equation*}
and
\begin{equation*}
\qvector_{j}=(\xmatrix_{\selectedfeatures,j})\transpose(\xmatrix_{\selectedfeatures}(\xmatrix_{\selectedfeatures})\transpose+\regparam\idmatrix)^{-1}\xmatrix_{\selectedfeatures,j},
\end{equation*}
provided that the vectors $\bm{f}$ and $\qvector$ are computed and stored in memory. The time complexity of computing $\bm{f}$ and $\qvector$ is $O(\arrowvert\selectedfeatures\arrowvert^3+\arrowvert\selectedfeatures\arrowvert^2\tsize)$, which is the same as that of training RLS via (\ref{firstform}). Alternatively, the constant time LOO predictions can be expressed in terms of the dual variables $\dualvec$ and the diagonal entries of the matrix $\dsikm$ as follows:
\begin{equation}\label{fastdualloo}
\labelvec_{j} - (\dsikm_{j,j})^{-1}\dualvec_{j}.
\end{equation}
Here, the time complexity of computing $\dualvec$ and the diagonal entries of the matrix $\dsikm$ is $O(\tsize^3+\tsize^2\arrowvert\selectedfeatures\arrowvert)$, which is equal to the complexity of training RLS via (\ref{secondform}). Thus, using either (\ref{fastprimalloo}) or (\ref{fastdualloo}), the LOO performance for the whole training set can be computed in $O(\min\{\arrowvert\selectedfeatures\arrowvert^2\tsize,\tsize^2\arrowvert\selectedfeatures\arrowvert\})$ time, which is equal to the time complexity of training RLS.

\section{Feature Selection}

In this section, we consider algorithmic implementations of feature selection for RLS with leave-one-out (LOO) criterion. We start by considering a straightforward wrapper approach which uses RLS as a black-box method in Section~\ref{wrappersection}. Next, we recall a previously proposed algorithm which provides results that are equivalent to those of the wrapper approach but which has computational short-cuts to speed up the feature selection in Section~\ref{ojedasection}. In Section~\ref{linearsection}, we present greedy RLS, our novel linear time algorithm, which again provides the same results as the wrapper approach but which has much lower computational and memory complexity than the previously proposed algorithms.

\subsection{Wrapper Approach}\label{wrappersection}

\begin{algorithm}
\label{wrapperalgo}
\KwIn{$\xmatrix\in\mathbb{R}^{\fsize\times\tsize}$, $\labelvec\in\mathbb{R}^{\tsize}$, $\desiredfcount$, $\regparam$}  
\KwOut{$\selectedfeatures$, $\primalvec$}
$\selectedfeatures\leftarrow\emptyset$\;
\While{$\arrowvert\selectedfeatures\arrowvert<\desiredfcount$}{
 $\bestloo\leftarrow\infty$\;
 $\bestfeaind\leftarrow 0$\;
 \ForEach{$i\in\{1,\ldots,\fsize\}\setminus\selectedfeatures$}{
  $\candidatefeaset\leftarrow\selectedfeatures\cup\{i\}$\;
  $\looerr\leftarrow 0$\;
  \ForEach{$j\in\{1,\ldots,\tsize\}$}{
   $\looindset\leftarrow\{1,\ldots,\tsize\}\setminus\{j\}$\;
   $\primalvec\leftarrow\trainalg(\xmatrix_{\candidatefeaset\looindset},\labelvec_{\looindset},\regparam)$\;
   $\pred\leftarrow\primalvec\transpose\xmatrix_{\candidatefeaset,j}$\;
   $\looerr\leftarrow\looerr + \lossfunction(\labelvec_{j},\pred)$\;
  }
  \If{$\looerr<\bestloo$}{
   $\bestloo\leftarrow\looerr$\;
   $\bestfeaind\leftarrow i$\;
  }
 }
 $\selectedfeatures\leftarrow\selectedfeatures\cup\{\bestfeaind\}$\;
}
$\primalvec\leftarrow\trainalg(\xmatrix_{\selectedfeatures},\labelvec,\regparam)$\;
\caption{Standard wrapper algorithm for RLS}
\end{algorithm}

Here, we consider the standard wrapper approach in which RLS is used as a black-box method which is re-trained for every feature set to be evaluated during the selection process and for every round of the LOO cross-validation. In forward selection, the set of all feature sets up to size $\desiredfcount$ is searched using hill climbing in the steepest descent direction (see e.g. \citet{russell1995ai}) starting from an empty set. The method is greedy in the sense that it adds one feature at a time to the set of selected features but no features are removed from the set at any stage. The wrapper approach is presented in Algorithm~\ref{wrapperalgo}. In the algorithm description, $\trainalg(\cdot,\cdot,\cdot)$ denotes the black-box training procedure for RLS which takes a data matrix, a label vector, and a value of the regularization parameter as input and returns a vector representation of the learned predictor $\primalvec$. The function $\lossfunction(\cdot,\cdot)$ computes the loss for one training example given its true label and a predicted label obtained from the LOO procedure. Typical losses are, for example, the squared loss for regression and zero-one error for classification.

Given that the computation of LOO performance requires $\tsize$ retrainings, that the forward selection goes through $O(\fsize)$ features in each iteration, and that $\desiredfcount$ features are chosen, the overall time complexity of the forward selection with LOO criterion is $O(\min\{\desiredfcount^3\tsize^2\fsize,\desiredfcount^2\tsize^3\fsize\})$. Thus, the wrapper approach is feasible with small training sets and in cases where the aim is to select only a small number of features. However, at the very least quadratic complexity with respect to both the size of the training set and the number of selected features make the standard wrapper approach impractical in large scale learning settings. The space complexity of the wrapper approach is $O(\fsize\tsize)$ which is equal to the cost of storing the data matrix $\xmatrix$.

An immediate reduction for the above considered computational complexities can be achieved via the short-cut methods for calculating the LOO performance presented in Section~\ref{RegFramework}. In machine learning and statistics literature, there have been several studies in which the computational short-cuts for LOO, as well as other types of cross-validation, have been used to speed up the evaluation of feature subset quality for non-regularized least-squares (see e.g \citet{shao1993linearcv,zhang1993mcv}) and for RLS (see e.g. \citet{ying2006fastloo,tang2006gene}). For the greedy forward selection, the short-cuts reduce the overall computational complexity to $O(\min\{\desiredfcount^3\tsize\fsize,\desiredfcount^2\tsize^2\fsize\})$, since LOO can then be calculated as efficiently as training the predictor itself. However, both \citet{ying2006fastloo} and \citet{tang2006gene} considered only the dual formulation (\ref{fastdualloo}) of the LOO speed-up which is expensive for large data sets. Thus, we are not aware of any studies in which the primal formulation (\ref{fastprimalloo}) would be implemented for the greedy forward selection for RLS.

\subsection{Previously Proposed Speed-Up}
\label{ojedasection}

We now present a feature selection algorithm proposed by \citet{ojeda2008lssvmselection} which the authors call low-rank updated LS-SVM. The features selected by the algorithm are equal to those by standard wrapper algorithm and by the method proposed by \citet{tang2006gene} but the method has a lower time complexity with respect to $\desiredfcount$ due to certain matrix calculus based computational short-cuts. The low-rank updated LS-SVM algorithm for feature selection is presented in Algorithm~\ref{ojedasalgo}.

\begin{algorithm}
\label{ojedasalgo}
\KwIn{$\xmatrix\in\mathbb{R}^{\fsize\times\tsize}$, $\labelvec\in\mathbb{R}^{\tsize}$, $\desiredfcount$, $\regparam$}  
\KwOut{$\selectedfeatures$, $\primalvec$}
$\selectedfeatures\leftarrow\emptyset$\;
$\dualvec\leftarrow\regparam^{-1}\labelvec$\;\label{dualinit}
$\dsikm\leftarrow\regparam^{-1}\idmatrix$\;\label{ginit}
\While{$\arrowvert\selectedfeatures\arrowvert<\desiredfcount$}{
 $\bestloo\leftarrow\infty$\;
 $\bestfeaind\leftarrow 0$\;
 \ForEach{$i\in\{1,\ldots,\fsize\}\setminus\selectedfeatures$}{\label{middleloop}
  $\rightvec \leftarrow (\xmatrix_{i})\transpose$\;
  $\widetilde{\dsikm}\leftarrow \dsikm-\dsikm\rightvec(1+\rightvec\transpose\dsikm\rightvec)^{-1}(\rightvec\transpose\dsikm)$\label{gupdone}\;
  $\tilde{\dualvec} \leftarrow \widetilde{\dsikm}\labelvec$\label{aupd1}\;
  $\looerr\leftarrow 0$\;
  \ForEach{$j\in\{1,\ldots,\tsize\}$}{
   $\pred\leftarrow\labelvec_{j} - (\widetilde{\dsikm}_{j,j})^{-1}\tilde{\dualvec}_{j}$\;
   $\looerr\leftarrow\looerr + \lossfunction(\labelvec_{j},\pred)$\;
  }
  \If{$\looerr<\bestloo$}{
   $\bestloo\leftarrow\looerr$\;
   $\bestfeaind\leftarrow i$\;
  }
 }
 $\rightvec \leftarrow (\xmatrix_{\bestfeaind})\transpose$\;
 $\dsikm\leftarrow \dsikm-\dsikm\rightvec(1+\rightvec\transpose\dsikm\rightvec)^{-1}(\rightvec\transpose\dsikm)$\;\label{gupdtwo}
 $\dualvec \leftarrow \dsikm\labelvec$\;\label{aupd2}
 $\selectedfeatures\leftarrow\selectedfeatures\cup\{\bestfeaind\}$\;
}
$\primalvec\leftarrow\xmatrix_{\selectedfeatures}\dualvec$\;
\caption{Low-rank updated LS-SVM}
\end{algorithm}

The low-rank updated LS-SVM is based on updating the matrix $\dsikm$ and the vector $\dualvec$ as new features are added into $\selectedfeatures$.
In the beginning of the forward update algorithm, the set $\selectedfeatures$ of selected features is empty. This means that the matrix $\dsikm$ contains no information of the data matrix $\xmatrix$. Instead, every entry of the matrix $\kernelm$ is equal to zero, and hence $\dsikm$ and $\dualvec$ are initialized to $\regparam^{-1}\idmatrix$ and $\regparam^{-1}\labelvec$, respectively, in lines \ref{dualinit} and \ref{ginit} in Algorithm~\ref{ojedasalgo}.
When a new feature candidate is evaluated, the LOO performance corresponding to the updated feature set can be calculated using the temporarily updated $\dsikm$ and $\dualvec$, denoted in the algorithm as $\widetilde{\dsikm}$ and $\tilde{\dualvec}$, respectively.
Here, the improved speed is based on two short-cuts. Namely, updating the matrix $\dsikm$ via the well known Sherman-Morrison-Woodbury (SMW) formula and using the fast LOO computation.

Let us first consider updating the matrix $\dsikm$. Formally, if the $i$th feature is to be evaluated, where $i\notin\selectedfeatures$, the algorithm calculates the matrix $\widetilde{\dsikm}$ corresponding the feature index set $\selectedfeatures+\{i\}$. According to (\ref{gmatrixdef}), the matrix $\widetilde{\dsikm}$ is
\begin{equation}\label{gupdslooow}
(\kernelm+\rightvec\rightvec\transpose+\regparam\idmatrix)^{-1},
\end{equation}
where $\kernelm$ is the kernel matrix defined as in (\ref{fullkernelmatrix}) without having the $i$th feature in $\selectedfeatures$ and $\rightvec=(\xmatrix_{i})\transpose$ contains the values of the $i$th feature for the training data. The time needed to compute (\ref{gupdslooow}) is cubic in $\tsize$, which provides no benefit compared to the standard wrapper approach. However, provided that the matrix $\dsikm$ is stored in memory, it is possible to speed up the computation of $\widetilde{\dsikm}$ via the SMW formula. Formally, the matrix $\widetilde{\dsikm}$ can also be obtained from
\begin{equation}\label{gupdate}
\dsikm-\dsikm\rightvec(1+\rightvec\transpose\dsikm\rightvec)^{-1}(\rightvec\transpose\dsikm)
\end{equation}
which requires a time quadratic in $\tsize$ provided that the multiplications are performed in the order indicated by the parentheses. Having calculated $\widetilde{\dsikm}$, the updated vector of dual variables $\tilde{\dualvec}$ can be obtained from
\begin{equation}\label{dualslow}
\widetilde{\dsikm}\labelvec
\end{equation}
also with $O(\tsize^2)$ floating point operations.

Now let us consider the efficient evaluation of the features with LOO performance. Provided that $\widetilde{\dsikm}$ and $\tilde{\dualvec}$ are already stored in memory, the LOO performance of RLS trained with the $\selectedfeatures+\{i\}$ training examples can be calculated via (\ref{fastdualloo}). After the feature, whose inclusion would be the most favorable with respect to the LOO performance, is found, it is added to the set of selected features and RLS solution is updated accordingly in the last four lines of the algorithm.

The outermost loop is run $\desiredfcount$ times, $\desiredfcount$ denoting the number of features to be selected. For each round of the outer loop, the middle loop is run $O(\fsize)$ times, because each of the $\fsize-\arrowvert\selectedfeatures\arrowvert$ features is evaluated in turn before the best of them is added to the set of selected features $\selectedfeatures$. Due to the efficient formula (\ref{fastdualloo}) for LOO computation, the calculation of the LOO predictions for the $\tsize$ requires only $O(\tsize)$ floating point operations in each run of the innermost loop. The complexity of the middle loop is dominated by the $O(\tsize^2)$ time needed for calculating $\widetilde{\dsikm}$ in line~\ref{gupdone} and $\tilde{\dualvec}$ in line~\ref{aupd1}, and hence the overall time complexity of the low-rank updated LS-SVM algorithm is $O(\desiredfcount\fsize\tsize^2)$.

The complexity with respect to $\desiredfcount$ is only linear which is much better than that of the standard wrapper approach, and hence the selection of large feature sets is made possible. However, feature selection with large training sets is still infeasible because of the quadratic complexity with respect to $\tsize$. In fact, the running time of the standard wrapper approach with the LOO short-cut can be shorter than that of the low-rank updated LS-SVM method in cases where the training set is large and the number of features to be selected is small.

The space complexity of the low-rank updated LS-SVM algorithm is $O(\fsize\tsize+\tsize^2)$, because the matrices $\xmatrix$ and $\dsikm$ require $O(\fsize\tsize)$ and $O(\tsize^2)$ space, respectively. Due to the quadratic dependence of $\tsize$, this space complexity is worse than that of the standard wrapper approach.

\subsection{Linear Time Forward Selection Algorithm}\label{linearsection}

Here, we present our novel algorithm for greedy forward selection for RLS with LOO criterion. We refer to our algorithm as greedy RLS, since in addition to feature selection point of view, it can also be considered as a greedy algorithm for learning sparse RLS predictors. Our method resembles the low-rank updated LS-SVM in the sense that it also operates by iteratively updating the vector of dual variables, and hence we define it on the grounds of the considerations of Section~\ref{ojedasection}. Pseudo code of greedy RLS is presented in Algorithm~\ref{ffsrls}.

The time and space complexities of the low-rank updated LS-SVM algorithm are quadratic in $\tsize$, because it involves updating the matrix $\dsikm$ (lines \ref{gupdone} and \ref{gupdtwo} in Algorithm~\ref{ojedasalgo}) and the vector $\dualvec$ (lines \ref{aupd1} and \ref{aupd2} in Algorithm~\ref{ojedasalgo}). In order to improve the efficiency of the algorithm by getting rid of the quadratic dependence of $\tsize$ in both space and time, we must avoid the explicit calculation and storing of the matrices $\dsikm$ and $\widetilde{\dsikm}$. Next, we present an improvement which solves these problems.

In the middle loop of the low-rank updated LS-SVM algorithm in Algorithm~\ref{ojedasalgo}, it is assumed that the matrix $\dsikm$ constructed according to the current set $\selectedfeatures$ of selected features is stored in memory. It is temporarily updated to matrix $\widetilde{\dsikm}$ corresponding to the set $\selectedfeatures+\{i\}$, where $i$ is the index of the feature to be evaluated. We observe that the LOO computations via the formula (\ref{fastdualloo}) require the vector of dual variables $\tilde{\dualvec}$ corresponding to $\selectedfeatures+\{i\}$ but only the diagonal elements of $\widetilde{\dsikm}$.

Let us first consider speeding up the computation of $\tilde{\dualvec}$. Since we already have $\dualvec$ stored in memory, we can take advantage of it when computing $\tilde{\dualvec}$. That is, since the value of the vector $\dualvec$ before the update is $\dsikm\labelvec$, its updated value $\tilde{\dualvec}$ can be, according to (\ref{gupdate}) and (\ref{dualslow}), obtained from
\begin{equation}\label{dualstillslow}
  \dualvec-\dsikm\rightvec(1+\rightvec\transpose\dsikm\rightvec)^{-1}(\rightvec\transpose\dualvec),
\end{equation}
where $\rightvec=(\xmatrix_{i})\transpose$. This does not yet help us much, because the form (\ref{dualstillslow}) still contains the matrix $\dsikm$ explicitly. However, if we assume that, in addition to $\dualvec$, the product $\dsikm\rightvec$ is calculated in advance and the result is stored in memory, calculating (\ref{dualstillslow}) only requires $O(\tsize)$ time. In fact, since the operation (\ref{dualstillslow}) is performed for almost every feature during each round of the greedy forward selection, we assume that we have an $\tsize\times\fsize$ cache matrix, denoted by
\begin{equation}\label{cachedef}
\cachematrix=\dsikm\xmatrix\transpose,
\end{equation}
computed in advance containing the required products for all features, each column corresponding to one product.

Now, given that $\cachematrix$ is available in memory, we calculate, in $O(\tsize)$ time, a vector
\begin{equation}\label{lvecform}
\leftvec=\cachematrix_{:,i}(1 + \rightvec\transpose \cachematrix_{:,i})^{-1}
\end{equation}
and store it in memory for later usage. This is done in lines \ref{uupdfirst} and \ref{uupdsecond} of the algorithm in Algorithm~\ref{ffsrls}. Consequently, (\ref{dualstillslow}) can be rewritten as
\begin{equation}\label{dualupdatefast}
\dualvec - \leftvec (\rightvec\transpose \dualvec).
\end{equation}
and hence the vector $\dualvec$ can be updated in $O(\tsize)$ time.

We also have to ensure that fast computation of the LOO performance. The use of (\ref{fastdualloo}) for computing the LOO for the updated feature set requires the diagonal elements of the matrix $\tilde{\dsikm}$. Let $\diagvec$ and $\tilde{\diagvec}$ denote the vectors containing the diagonal elements of $\dsikm$ and $\tilde{\dsikm}$, respectively. We observe that the SMW formula (\ref{gupdate}) for calculating $\tilde{\dsikm}$ can be rewritten as
\begin{equation}\label{gupd2}
\dsikm-\leftvec(\cachematrix_{:,i})\transpose.
\end{equation}
Now, we make an assumption that, instead of having the whole $\dsikm$ being stored in memory, we have only stored its diagonal elements, that is, the vector $\diagvec$. Then, according to (\ref{gupd2}), the $j$th element of the vector $\tilde{\diagvec}$ can be computed from
\begin{equation}\label{diagelemupdate}
  \diagvec_{j} - \leftvec_{j}\cachematrix_{j,i}
\end{equation}
requiring only a constant time, the overall time needed for computing $\tilde{\diagvec}$ again becoming $O(\tsize)$.

Thus, provided that we have all the necessary caches available, evaluating each feature requires $O(\tsize)$ time, and hence one pass through the whole set of $\fsize$ features needs $O(\tsize\fsize)$ floating point operations. But we still have to ensure that the caches can be initialized and updated efficiently enough.

In the initialization phase of the greedy RLS algorithm (lines~\ref{grlsinitbegin}-\ref{grlsinitend} in Algorithm~\ref{ffsrls}) the set of selected features is empty, and hence the values of $\dualvec$, $\diagvec$, and $\cachematrix$ are initialized to $\regparam^{-1}\labelvec$, $\regparam^{-1}\bm{1}$, and $\regparam^{-1}\xmatrix\transpose$, respectively, where $\bm{1}\in\mathbb{R}^{\tsize}$ is a vector having every entry equal to $1$. The computational complexity of the initialization phase is dominated by the $O(\tsize\fsize)$ time required for initializing $\cachematrix$. Thus, the initialization phase is no more complex than one pass through the features.

When a new feature is added into the set of selected features, the vector $\dualvec$ is updated according to (\ref{dualupdatefast}) and the vector $\diagvec$ according to (\ref{diagelemupdate}). Putting together (\ref{cachedef}), (\ref{lvecform}), and (\ref{gupd2}), the cache matrix $\cachematrix$ can be updated via
\begin{equation*}
\cachematrix-\leftvec(\rightvec\transpose\cachematrix),
\end{equation*}
which requires $O(\tsize\fsize)$ time. This is equally expensive as the above introduced fast approach for trying each feature at a time using LOO as a selection criterion.

Finally, if we are to select altogether $\desiredfcount$ features, the overall time complexity of greedy RLS becomes $O(\desiredfcount\tsize\fsize)$. The space complexity is dominated by the matrices $\xmatrix$ and $\cachematrix$ which both require $O(\tsize\fsize)$ space.

\begin{algorithm}
\label{ffsrls}
\KwIn{$\xmatrix\in\mathbb{R}^{\fsize\times\tsize}$, $\labelvec\in\mathbb{R}^{\tsize}$, $\desiredfcount$, $\regparam$}  
\KwOut{$\selectedfeatures$, $\primalvec$}
$\dualvec\leftarrow\regparam^{-1}\labelvec$\label{grlsinitbegin}\;
$\diagvec\leftarrow\regparam^{-1}\bm{1}$\;
$\cachematrix\leftarrow\regparam^{-1}\xmatrix\transpose$\;
$\selectedfeatures\leftarrow\emptyset$\label{grlsinitend}\;
\While{$\arrowvert\selectedfeatures\arrowvert<\desiredfcount$}{
 $\bestloo\leftarrow\infty$\;
 $\bestfeaind\leftarrow 0$\;
 \ForEach{$i\in\{1,\ldots,\fsize\}\setminus\selectedfeatures$}{
  $\rightvec \leftarrow (\xmatrix_{i})\transpose$\;
  $\leftvec \leftarrow \cachematrix_{:,i} (1 + \rightvec \cachematrix_{:,i})^{-1}$\label{uupdfirst}\;
  $\tilde{\dualvec} \leftarrow \dualvec - \leftvec (\rightvec\transpose \dualvec)$\;
  $\looerr\leftarrow 0$\;
  \ForEach{$j\in\{1,\ldots,\tsize\}$}{
   $\tilde{\diagvec}_{j} \leftarrow \diagvec_{j} - \leftvec_{j}\cachematrix_{j,i}$\;
   $\pred\leftarrow\labelvec_{j} - (\tilde{\diagvec}_{j})^{-1}\tilde{\dualvec}_{j}$\;
   $\looerr\leftarrow\looerr + \lossfunction(\labelvec_{j},\pred)$\;
  }
  \If{$\looerr<\bestloo$}{
   $\bestloo\leftarrow\looerr$\;
   $\bestfeaind\leftarrow i$\;
  }
 }
 $\rightvec \leftarrow (\xmatrix_{\bestfeaind})\transpose$\;
 $\leftvec \leftarrow \cachematrix_{:,\bestfeaind} (1 + \rightvec\transpose \cachematrix_{:,\bestfeaind})^{-1}$\label{uupdsecond}\;
 $\dualvec \leftarrow \dualvec - \leftvec (\rightvec\transpose \dualvec)$\;
 \ForEach{$j\in\{1,\ldots,\tsize\}$}{
  $\diagvec_{j} \leftarrow \diagvec_{j} - \leftvec_{j}\cachematrix_{j,\bestfeaind}$\;
 }
 $\cachematrix\leftarrow\cachematrix-\leftvec(\rightvec\transpose\cachematrix)$\;
 $\selectedfeatures\leftarrow\selectedfeatures\cup\{\bestfeaind\}$\;
}
$\primalvec\leftarrow\xmatrix_{\selectedfeatures}\dualvec$\;
\caption{Greedy RLS algorithm proposed by us}
\end{algorithm}

\section{Experimental results}

In Section~\ref{efficiencysection}, we explore the computational efficiency of the introduced
method and compare it with the best previously proposed implementation. In Section~\ref{qualitysection}, we explore the quality of the feature selection process.
In addition, we conduct measurements of the degree to which the LOO criterion overfits in Section~\ref{overfitsection}.

\subsection{Computational efficiency}\label{efficiencysection}

First, we present experimental results about the scalability of our method.
We use randomly generated data from two normal distributions with $1000$
features of which $50$ are selected. The number of examples is varied. In the first experiment we vary the
number of training examples between $500$ and $5000$, and provide a comparison
to the low-rank updated LS-SVM method \citet{ojeda2008lssvmselection}. In the
second experiment the training set size is varied between $1000$ and $50000$. We do not consider the baseline method in
the second experiment, as it does not scale up to the considered training
set sizes. The runtime experiments were run on a modern desktop computer
with 2.4 GHz Intel Core 2 Duo E6600 processor,
8 GB of main memory, and 64-bit Ubuntu Linux 9.10 operating system.

We note that the running times of these two methods are not
affected by the choice of the regularization parameter, or the distribution of
the features or the class labels. This is in contrast to iterative optimization
techniques commonly used to train for example support vector machines \cite{bottou2007svmsolvers}.
Thus we can draw general conclusions about the scalability of the methods from
experiments with a fixed value for the regularization parameter, and synthetic
data.

\begin{figure}
\includegraphics[width=\linewidth]{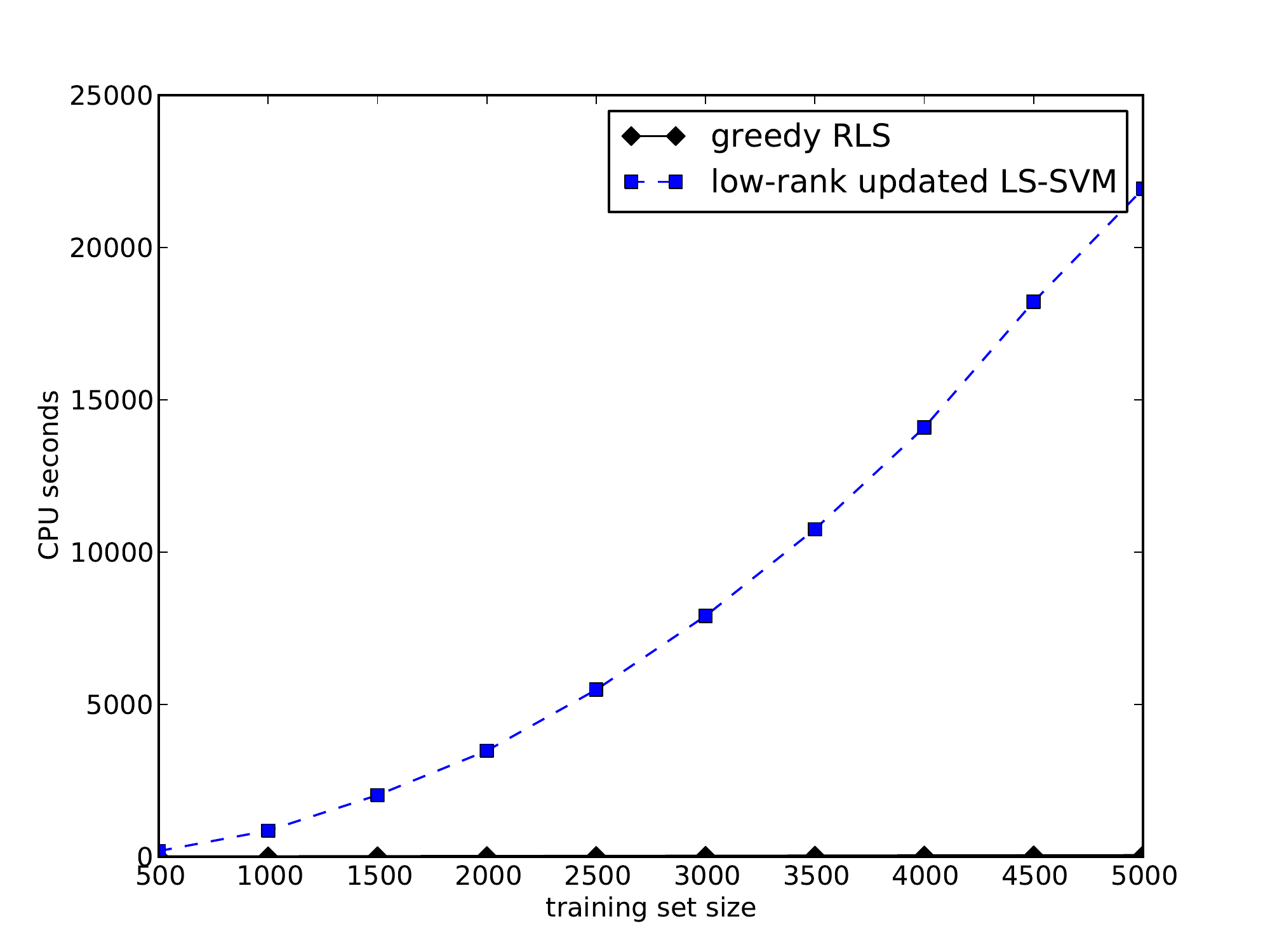}
\caption{Running times in CPU seconds for the proposed greedy RLS 
method and the and the low-rank updated LS-SVM 
of \citet{ojeda2008lssvmselection}. Linear scaling on y-axis.}
\label{fig:LinearComparison}
\end{figure}

\begin{figure}
\includegraphics[width=\linewidth]{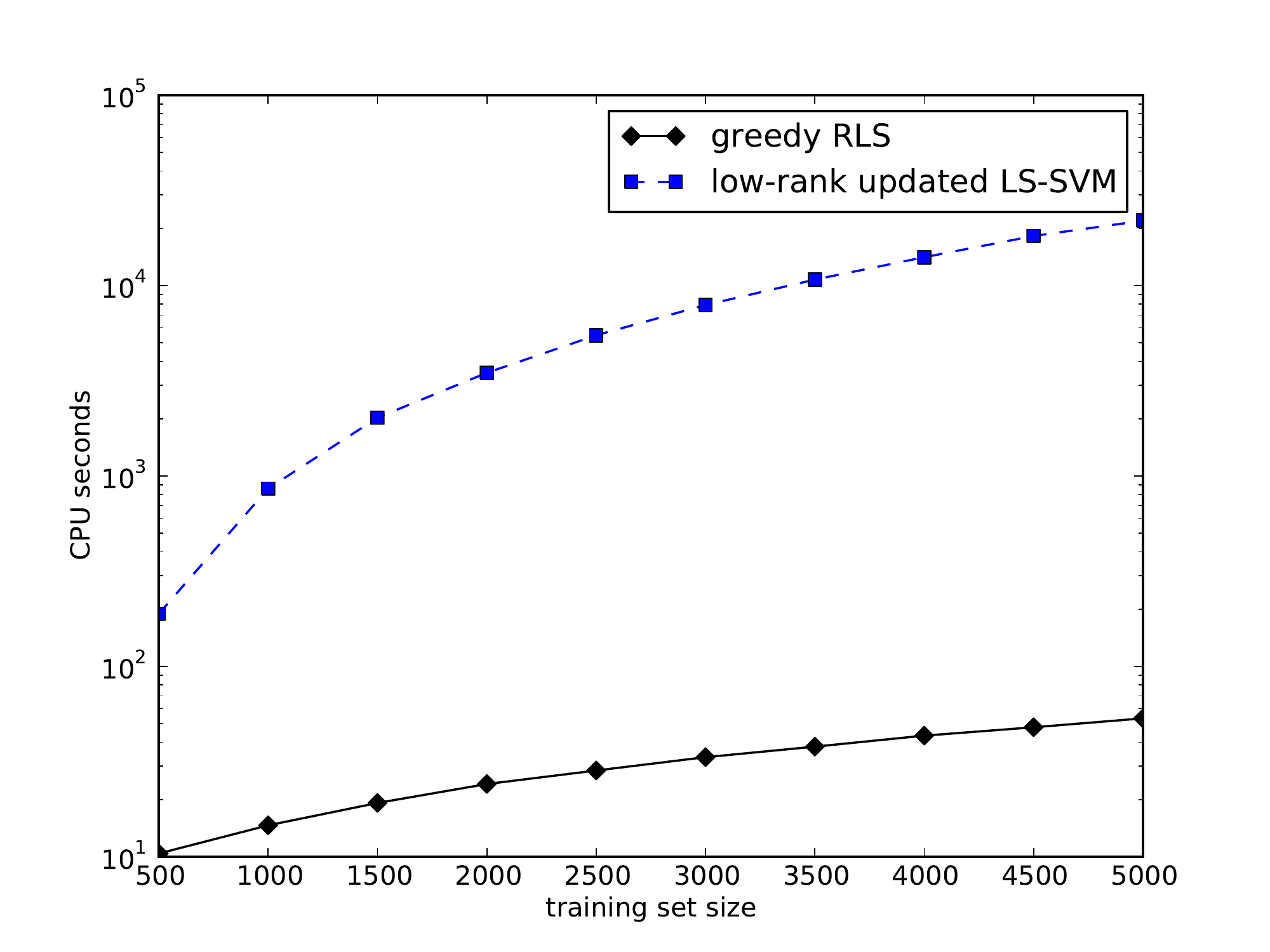}
\caption{Running times in CPU seconds for the proposed greedy RLS 
method and the and the low-rank updated LS-SVM 
of \citet{ojeda2008lssvmselection} Logarithmic scaling on y-axis.}
\label{fig:LogarithmicComparison}.
\end{figure}

\begin{figure}
\includegraphics[width=\linewidth]{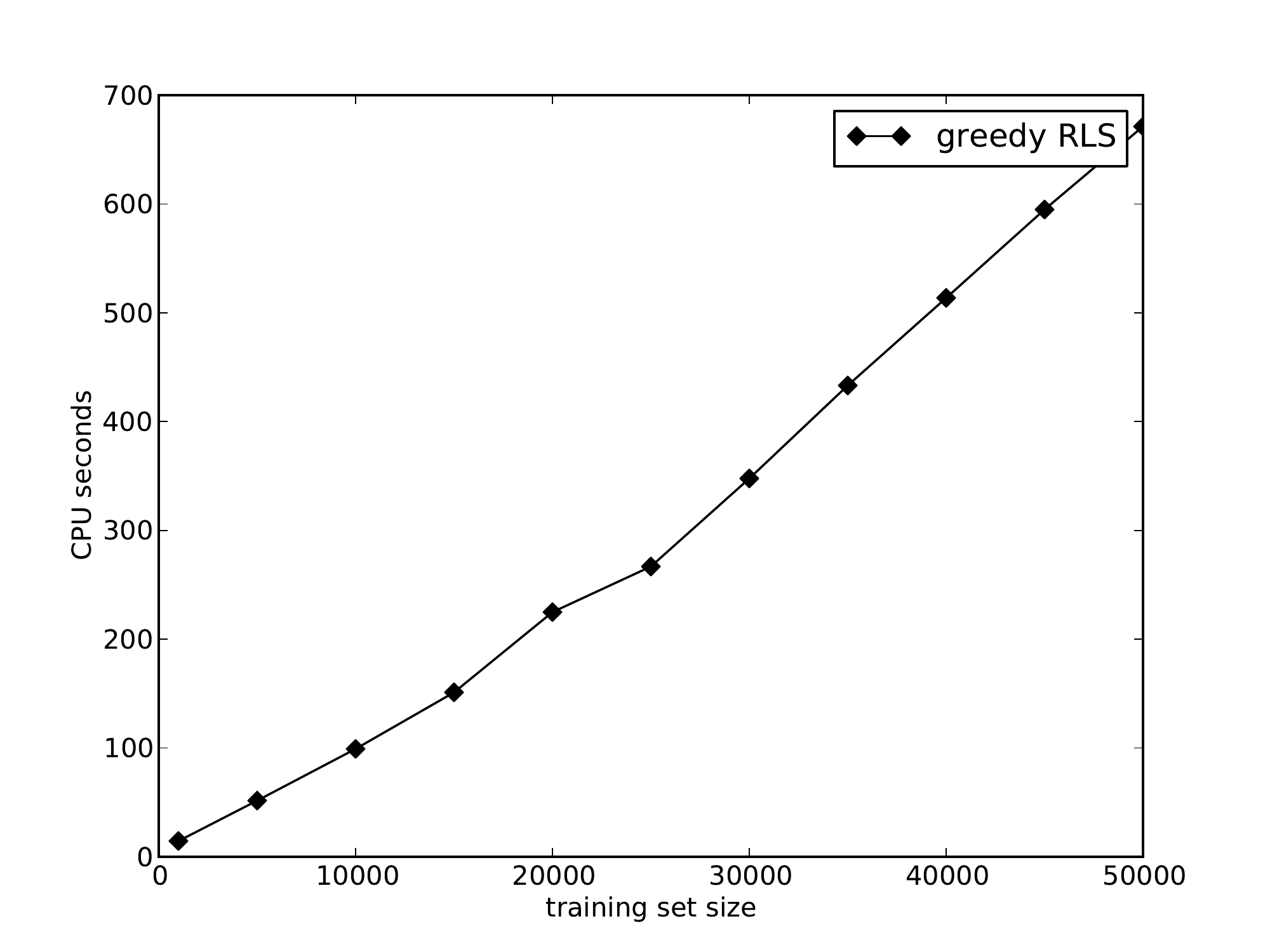}
\caption{Running times in CPU seconds for the proposed greedy RLS 
method.}
\label{fig:LscaleExperiment}
\end{figure}

In Fig.~\ref{fig:LinearComparison} are the runtime comparisons with linear
scaling on y-axis, and in Fig.~\ref{fig:LogarithmicComparison} with
logarithmic scaling. The results are consistent with the algorithmic complexity
analysis of the methods. The method of \citet{ojeda2008lssvmselection} shows quadratic scaling
with respect to the number of training examples, while the proposed method
scales linearly. In Fig.~\ref{fig:LscaleExperiment} are the running times for
the proposed method for up to 50000 training examples, in which case selecting
50 features out of 1000 took a bit less than twelve minutes.

\subsection{Quality of selected features}\label{qualitysection}

In this section, we explore the quality of the feature selection process.
We recall that our greedy RLS algorithm leads to equivalent results as the algorithms proposed by
\citet{tang2006gene} and by \citet{ojeda2008lssvmselection}, while being computationally much more
efficient. The aforementioned authors have in their work shown that the 
approach compares favorably to other commonly used feature selection techniques. However, due to computational constraints the earlier evaluations have been run
on small data sets consisting only of tens or hundreds of examples. We extend
the evaluation to large scale learning with the largest of the considered data
sets consisting of more than hundred thousand training examples. Wrapper
selection methods do not typically scale to such data set sizes, especially when using cross-validation as a selection criterion.
Thus we consider as a baseline an approach which chooses $\desiredfcount$
features at random. This is a good sanity-check, since training RLS with this
approach requires only $O(\min(\desiredfcount^2\tsize,\desiredfcount\tsize^2))$ time
that is even less than the time required by greedy RLS.

We consider the binary classification setting, the
quality of a selected feature set is measured by the classification
accuracy of a model trained on these features, evaluated on independent
test data. The experiments are run on a number of publicly available
benchmark data sets, whose characteristics are summarized in
Table~\ref{datasets}. The mnist dataset is originally a multi-class digit
recognition task, here we consider the binary classification task of recognizing
the digit 5. 

\begin{table}
\caption{Data sets}
\centering
\begin{tabular}{|l|l|l|}
\hline
data set & \#instances & \#features \\
\hline  
adult & 32561 & 123 \\
australian & 683 & 14 \\
colon-cancer & 62 & 2000 \\
german.numer & 1000 & 24 \\
icjnn1 & 141691 & 22 \\
mnist5 & 70000 & 780 \\
\hline
\end{tabular}
\centering
\label{datasets}
\end{table}

All the presented results are average accuracies over stratified
tenfold-cross validation run on the full data sets.
On each of the ten cross-validation rounds, before the feature
selection experiment is run we select the value of the regularization
parameter as follows. We train the learner on the training folds
using the full feature set, and perform a grid search to choose a
suitable regularization parameter value based on leave-one-out
performance.

Using the chosen regularization parameter value, we begin the
incremental feature selection process on the training folds.
One at a time we select the feature, whose addition to the model
provides highest LOO accuracy estimate on the training folds.
Each time a feature has been selected, the generalization performance
of the updated model is evaluated on the test fold. The
process is carried on until all the features have been selected.

\begin{figure}
\includegraphics[width=\linewidth]{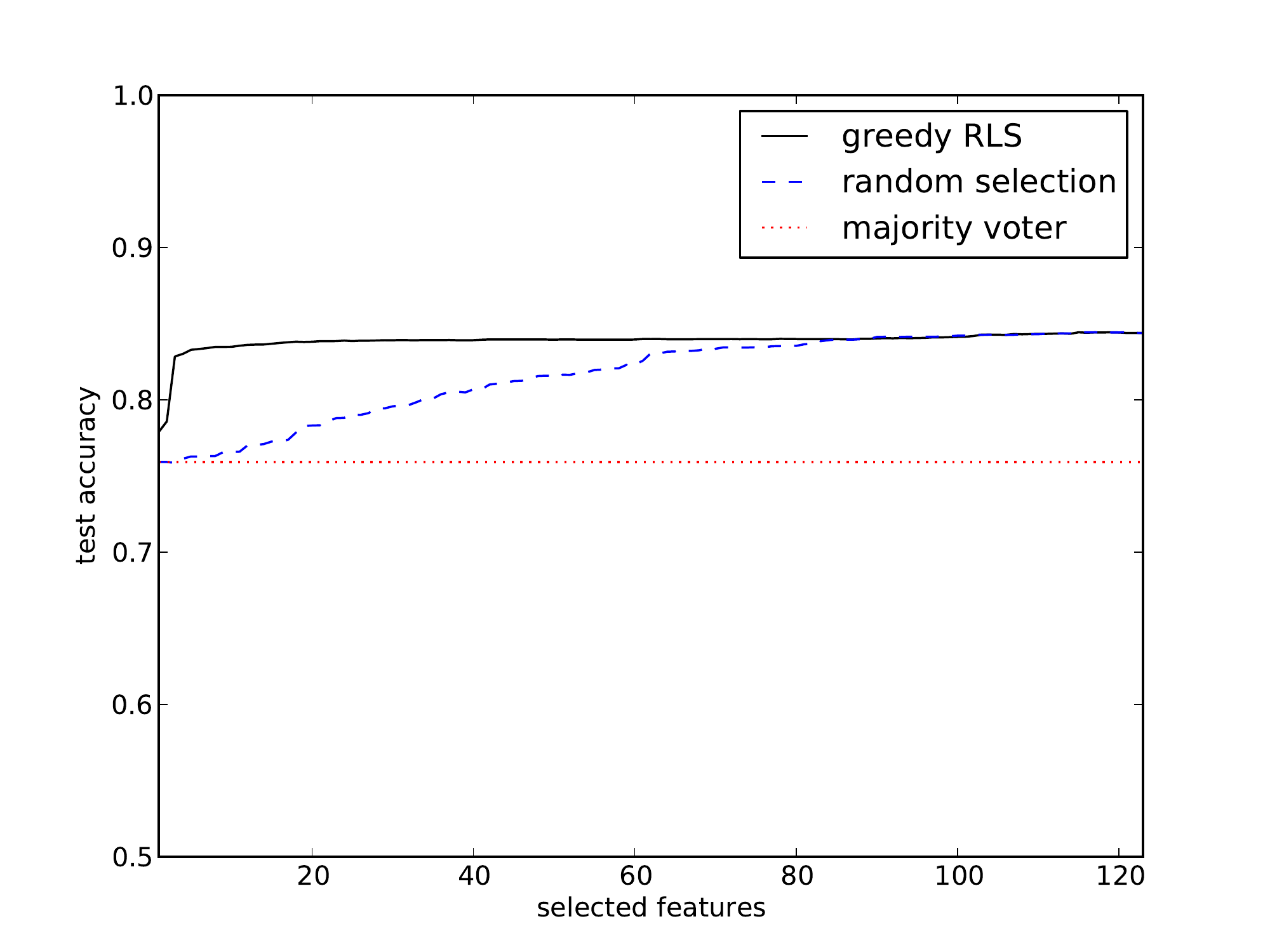}
\caption{Performance on the adult data set.}
\label{fig:adult}
\end{figure}

\begin{figure}
\includegraphics[width=\linewidth]{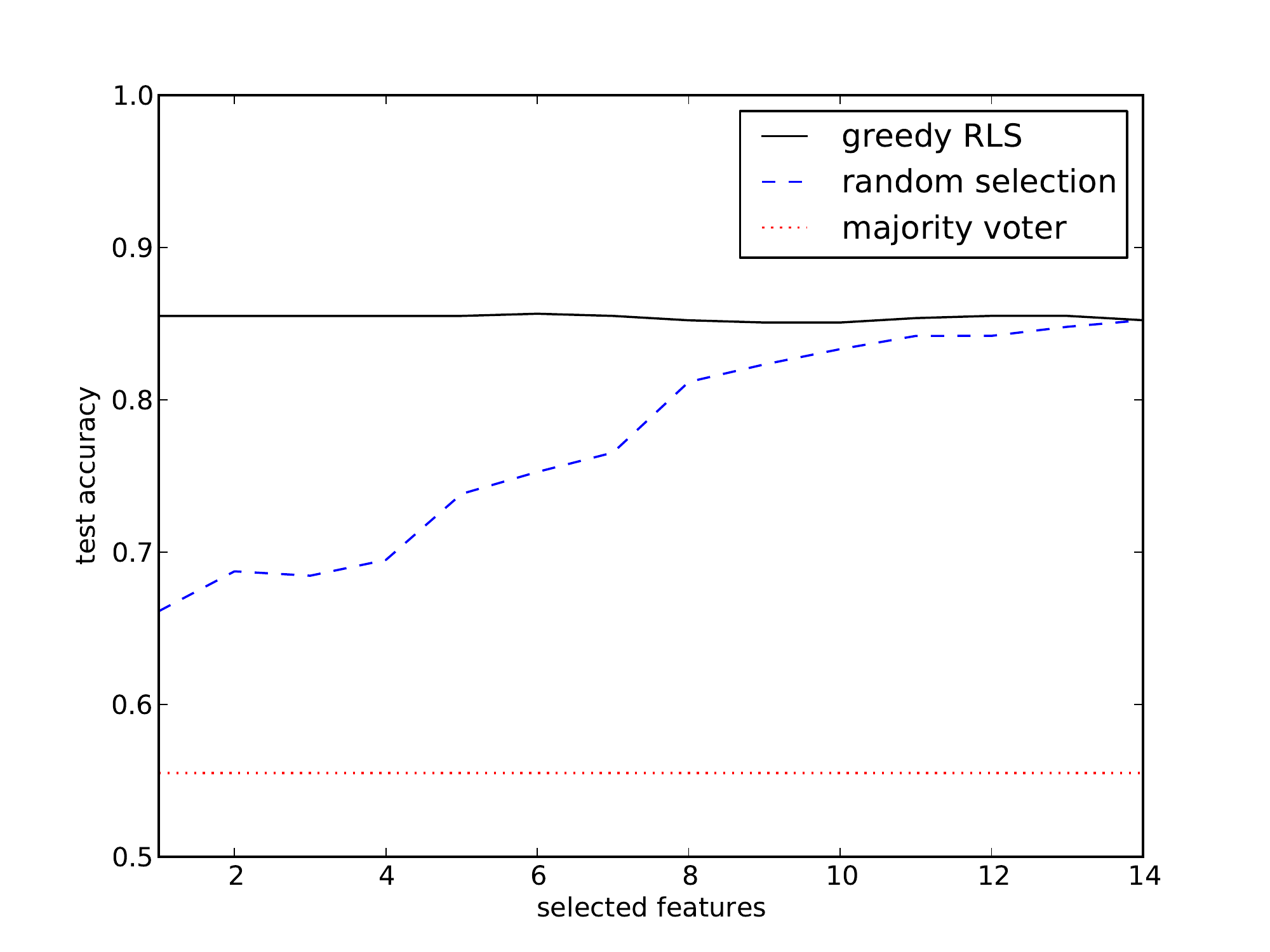}
\caption{Performance on the australian data set.}
\label{fig:australian}
\end{figure}

\begin{figure}
\includegraphics[width=\linewidth]{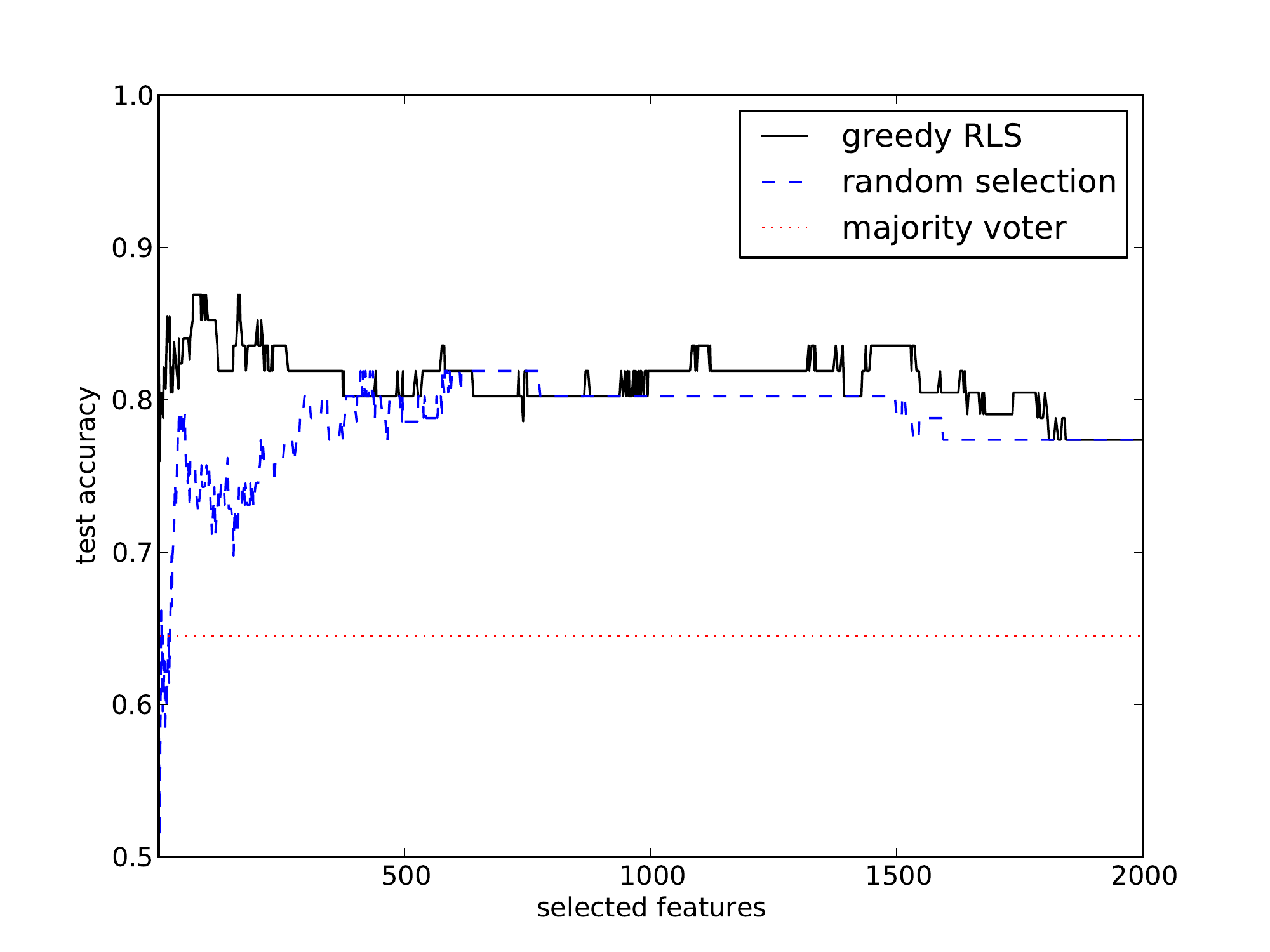}
\caption{Performance on the colon-cancer data set.}
\label{fig:colon-cancer}
\end{figure}

\begin{figure}
\includegraphics[width=\linewidth]{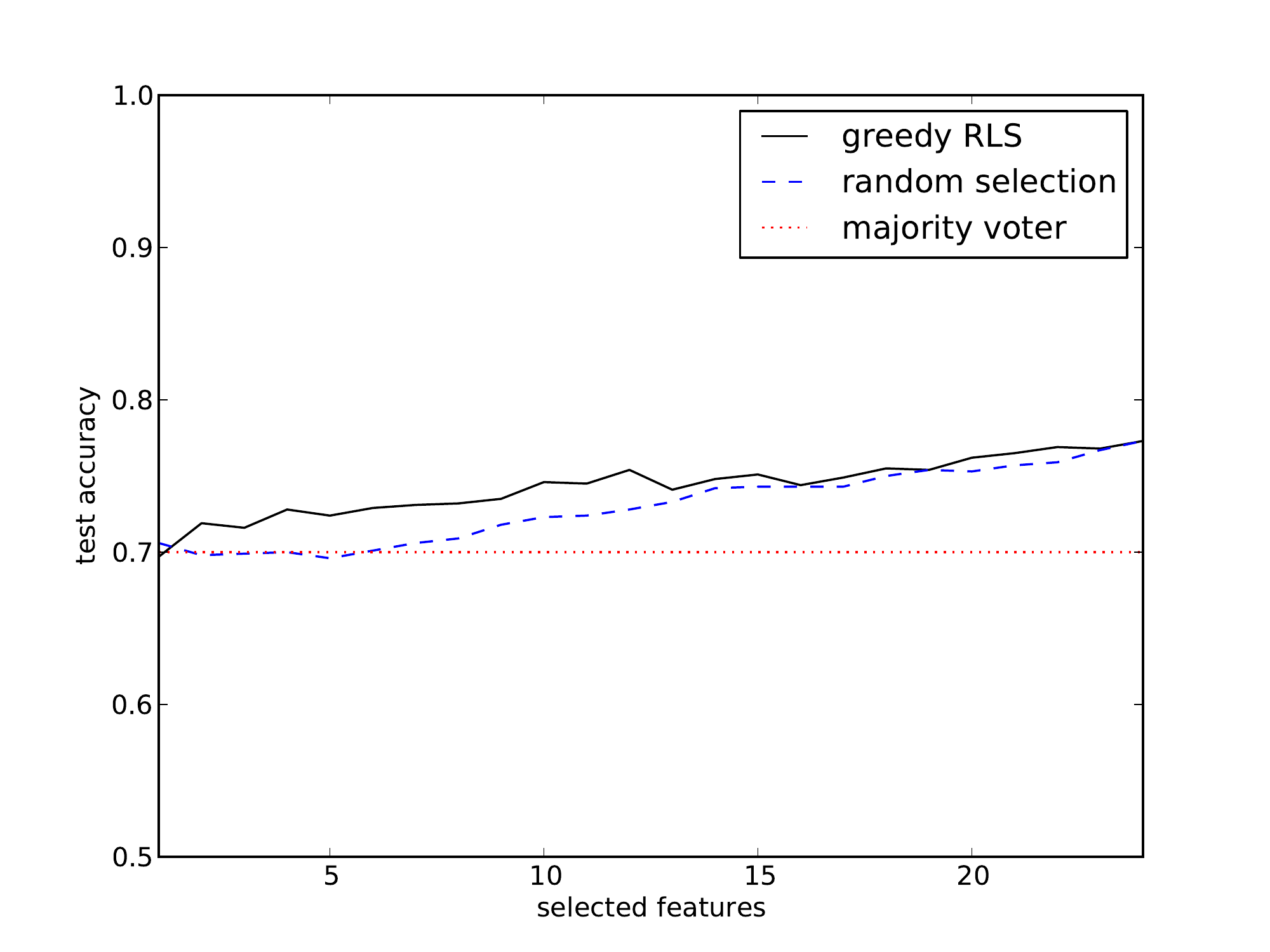}
\caption{Performance on the german.numer data set.}
\label{fig:german.numer}
\end{figure}

\begin{figure}
\includegraphics[width=\linewidth]{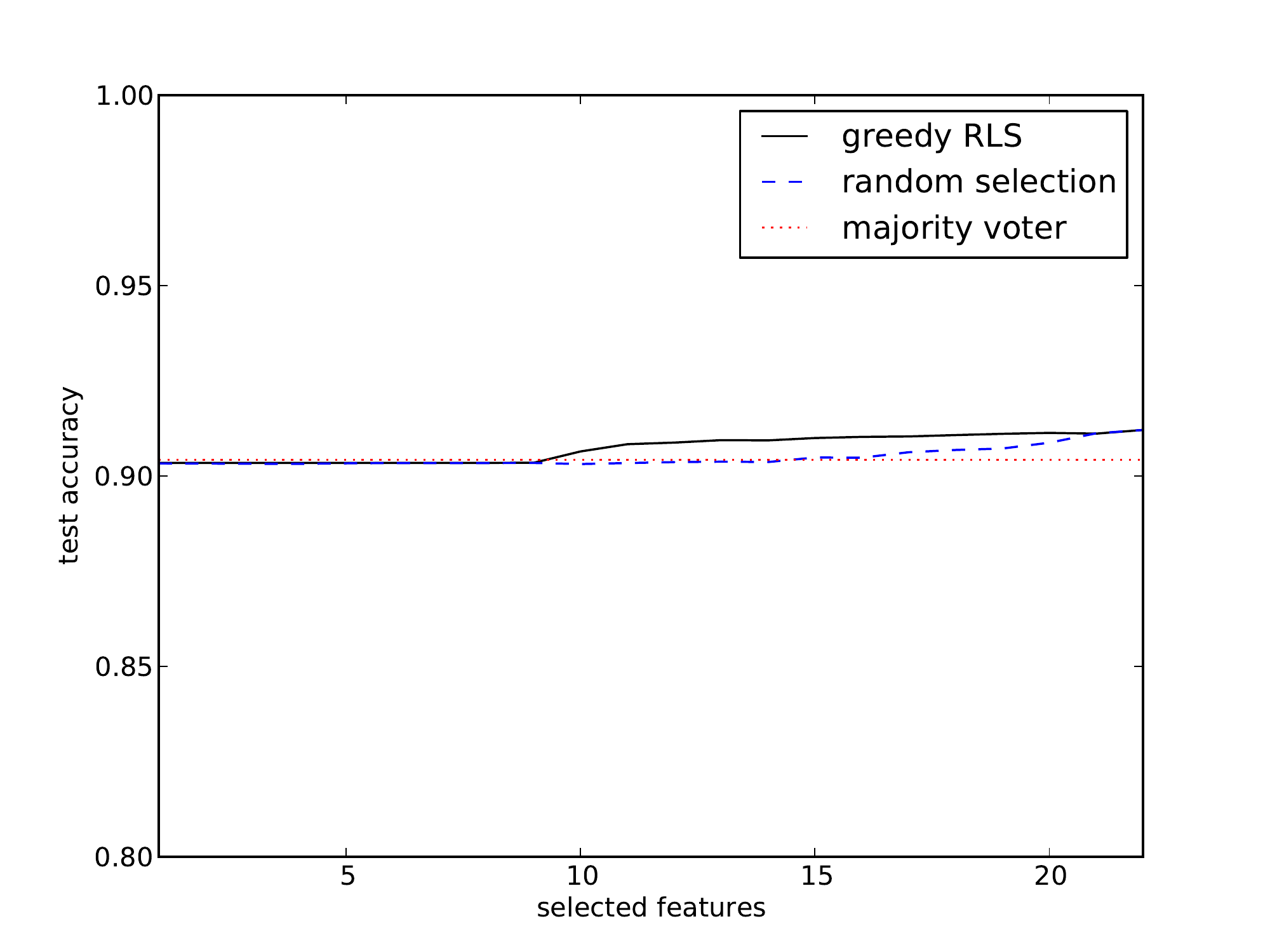}
\caption{Performance on the ijcnn1 data set.}
\label{fig:ijcnn1}
\end{figure}

\begin{figure}
\includegraphics[width=\linewidth]{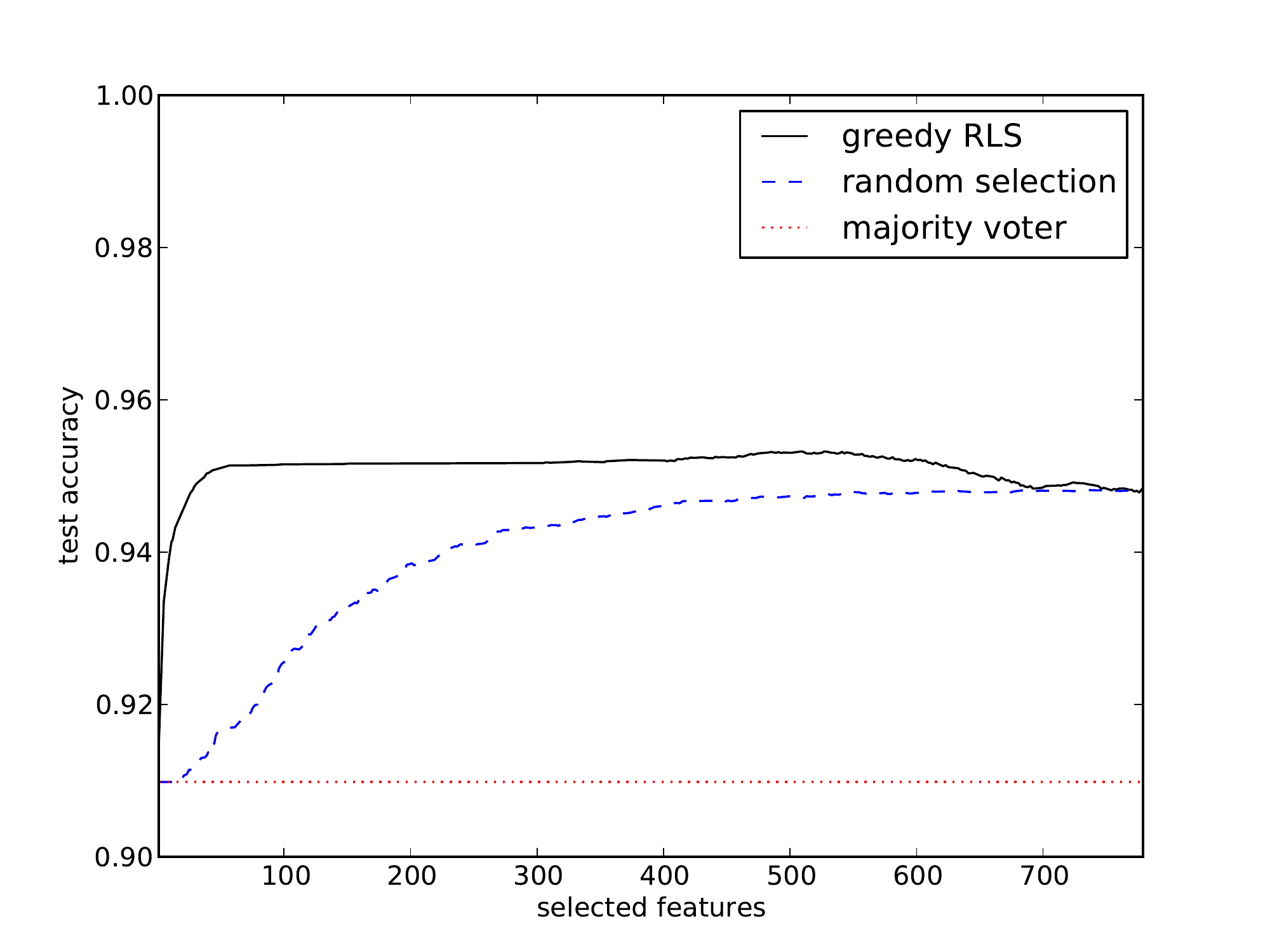}
\caption{Performance on the mnist5 data set.}
\label{fig:mnist}
\end{figure}

In the presented figures we plot the number of selected features
against the accuracy achieved on the test fold, averaged over
the ten cross-validation rounds.
In Fig.~\ref{fig:adult} are the results for the adult,
in Fig.~\ref{fig:australian} for the australian,
in Fig.~\ref{fig:colon-cancer} for colon-cancer,
in Fig.~\ref{fig:german.numer} for german.numer,
in Fig.~\ref{fig:ijcnn1} for ijcnn1, and
in Figure~\ref{fig:mnist} for the mnist5 data set.

On all data sets, the proposed approach
clearly outperforms the random selection strategy, suggesting
that the leave-one-out criterion leads to selecting informative
features. In most cases with only a small subset of features
one can achieve as good performance as with all the features.

\subsection{Overfitting in feature selection}\label{overfitsection}

One concern with using the LOO estimate for selecting
features is that the estimate may overfit to the training data.
If one considers large enough set of features, it is quite likely
that some features will by random chance lead to improved LOO estimate,
while not generalizing outside the training set. We explore to what extent
and in which settings this is a problem by comparing the LOO and test
accuracies in the previously introduced experimental setting. Again,
we plot the number of selected features against the accuracy estimates.

\begin{figure}
\includegraphics[width=\linewidth]{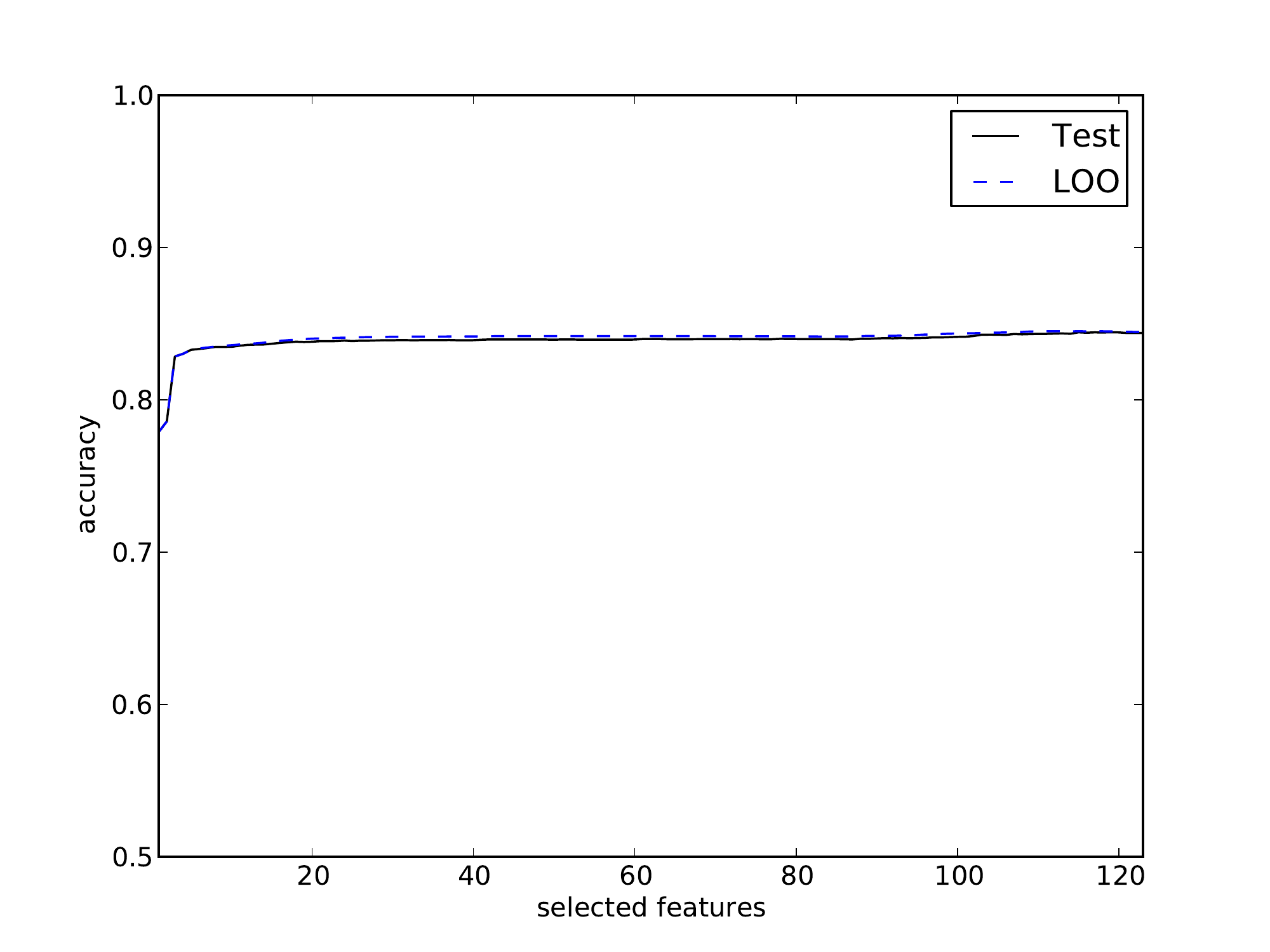}
\caption{Test vs. LOO accuracy on the adult data set.}
\label{fig:adult2}
\end{figure}

\begin{figure}
\includegraphics[width=\linewidth]{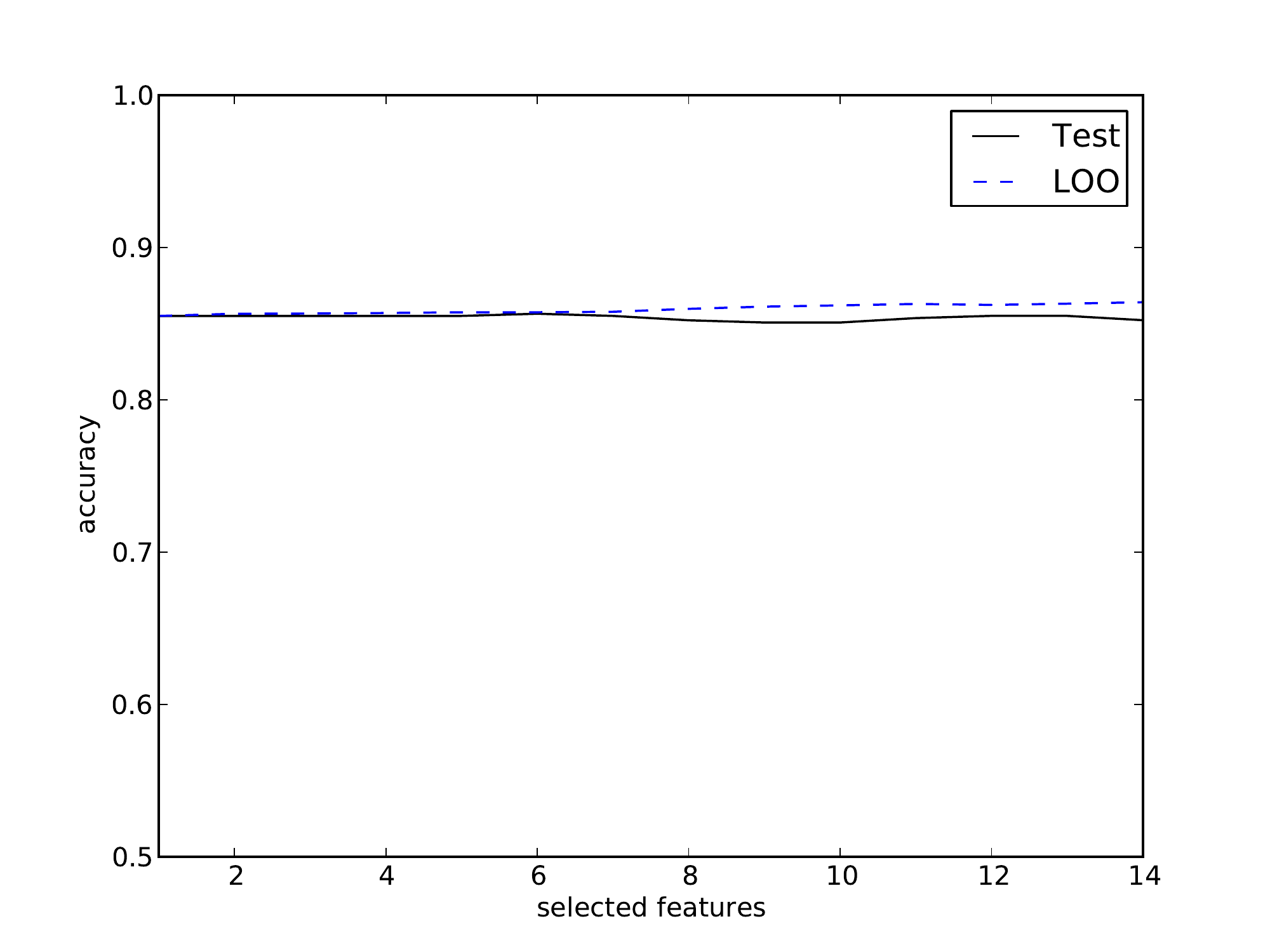}
\caption{Test vs. LOO accuracy on the australian data set.}
\label{fig:australian2}
\end{figure}

\begin{figure}
\includegraphics[width=\linewidth]{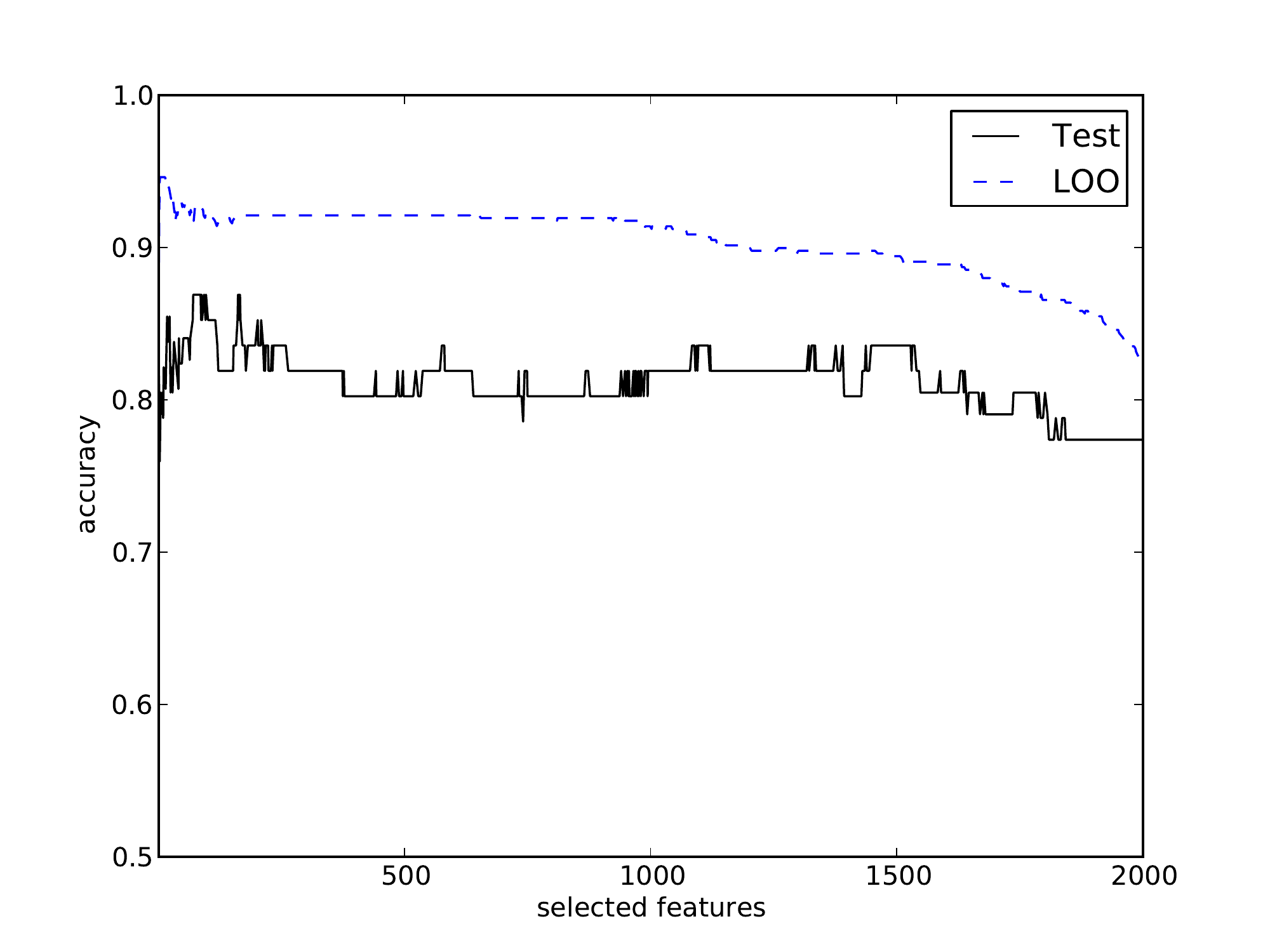}
\caption{Test vs. LOO accuracy on the colon-cancer data set.}
\label{fig:colon-cancer2}
\end{figure}

\begin{figure}
\includegraphics[width=\linewidth]{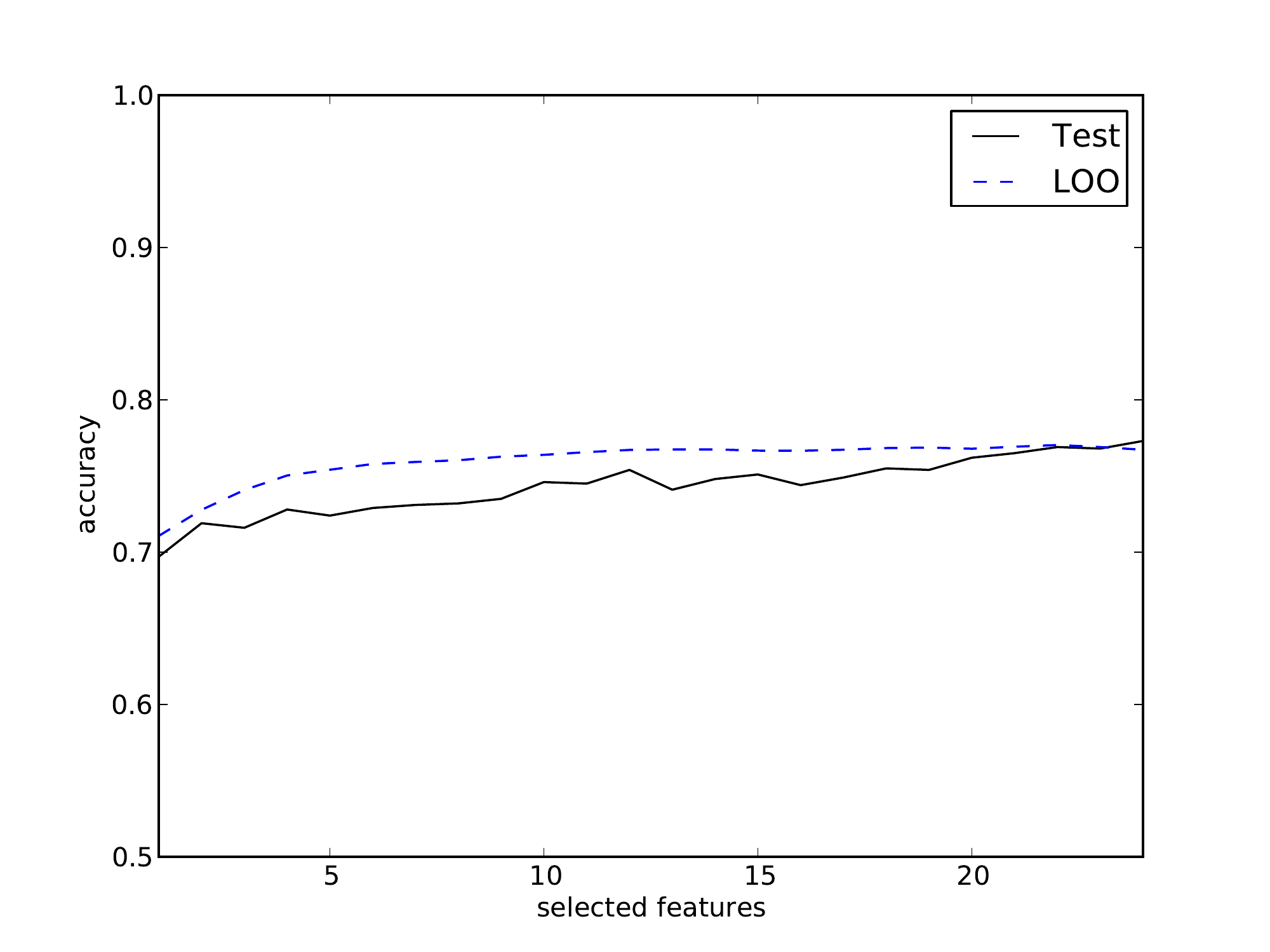}
\caption{Test vs. LOO accuracy on the german.numer data set.}
\label{fig:german.numer2}
\end{figure}

\begin{figure}
\includegraphics[width=\linewidth]{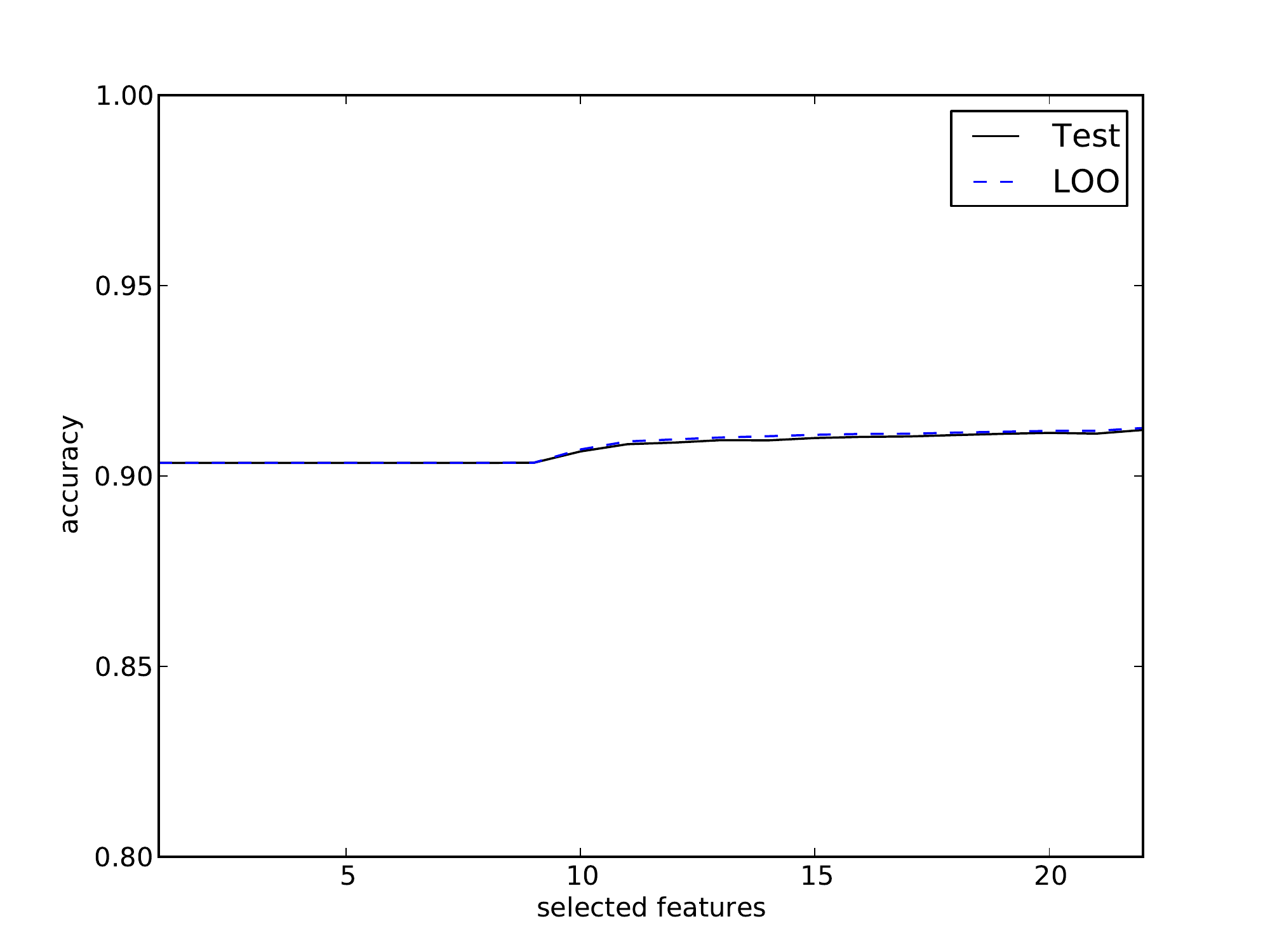}
\caption{Test vs. LOO accuracy on the ijcnn1 data set.}
\label{fig:ijcnn12}
\end{figure}

\begin{figure}
\includegraphics[width=\linewidth]{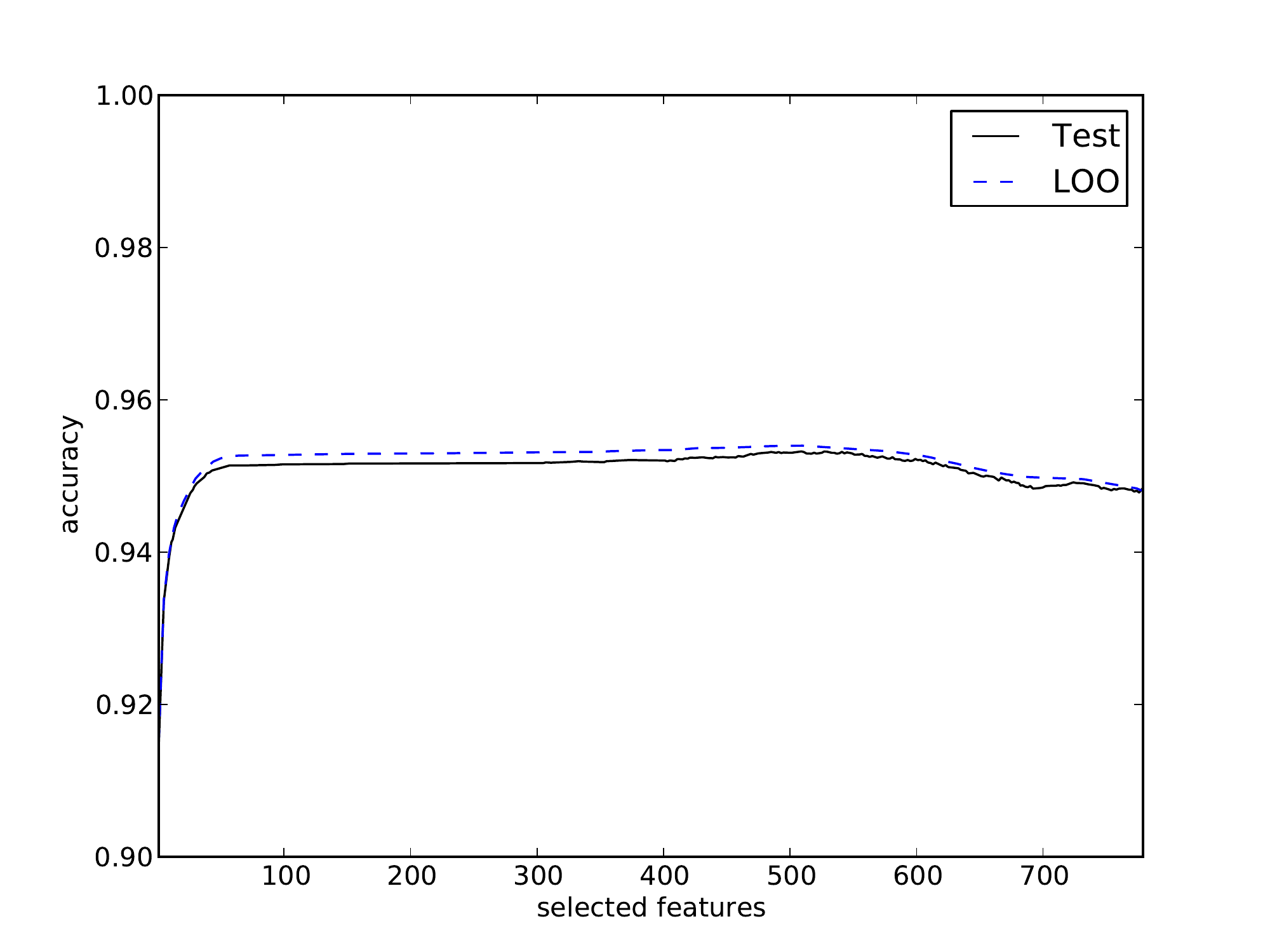}
\caption{Test vs. LOO accuracy on the mnist5 data set.}
\label{fig:mnist2}
\end{figure}

In most cases the LOO and test accuracies are close to identical
(Figures \ref{fig:adult2}, \ref{fig:australian2}, \ref{fig:ijcnn12},
and \ref{fig:mnist2}).
However, on the colon-cancer and german.numer data sets
(Figures \ref{fig:colon-cancer2} and
\ref{fig:german.numer2}) we see evidence of overfitting,
with LOO providing over-optimistic evaluation of performance.
The effect is especially clear
on the colon-cancer data set, which has $2000$ features, and only
$62$ examples. The results suggest that reliable feature selection
can be problematic on small high-dimensional data sets, the type
of which are often considered for example in bioinformatics. On
larger data sets the LOO estimate is quite reliable.

\section{Discussion and Future Directions}

In this paper, we present greedy RLS, a linear time algorithm for feature selection for RLS. The algorithm uses LOO as a criterion for evaluating the goodness of the feature subsets and greedy forward selection as a search strategy in the set of possible feature sets. The proposed algorithm opens several directions for future research. In this section, we will briefly discuss some of the directions which we consider to be the most promising ones.

Greedy RLS can quite straightforwardly be generalized to use different types of cross-validation criteria, such as $n$-fold or repeated $n$-fold. These are motivated by their smaller variance compared to the leave-one-out and they have also been shown to have better asymptotic convergence properties for feature subset selection for ordinary least-squares \citep{shao1993linearcv}. This generalization can be achieved by using the short-cut methods developed by us \citep{pahikkala06rls} and by \citet{an2007lssvmcv}.

In this paper, we focus on the greedy forward selection approach but an analogous method can also be constructed for backward elimination. However, backward elimination would be computationally much more expensive than forward selection, because an RLS predictor has to be trained first with the whole feature set. In the feature selection literature, approaches which combine both forward and backward steps, such as the floating search methods (see e.g. \citet{pudil1994floating,li2006piecewise}), have also been proposed. Recently, \citet{zhang2009adaptive} considered a modification of the forward selection for least-squares, which performs corrective steps instead of greedily adding a new feature in each iteration. The considered learning algorithm was ordinary least-squares and the feature subset selection criterion was empirical risk. The modified approach was shown to have approximately the same computational complexity (in number of iterations) but better performance than greedy forward selection or backward elimination. A rigorous theoretical justification was also presented for the modification. While this paper concentrates on the algorithmic implementation of the feature subset selection for RLS, we aim to investigate the performance of this type of modifications in our future work.

The computational short-cuts presented in this paper are possible due to the closed form solution the RLS learning algorithms has. Such short-cuts are also available for variations of RLS such as RankRLS, an RLS based algorithm for learning to rank proposed by us in \citet{pahikkala2007rankrls,pahikkala2009ranking}. In our future work, we also plan to design and implement similar feature selection algorithms for RankRLS.

Analogously to the feature selection methods, many approaches has been developed also for so-called reduced set selection used in context of kernel-based learning algorithms (see e.g \citet{smola00sparse,rifkin2003rls} and references therein). For example, incremental approaches for selecting this set for regularized least-squares has been developed by \citet{jiao2007fastsparse}. Closely related to the reduced set methods, we can also mention the methods traditionally used for selecting centers for radial basis function networks. Namely, the algorithms based on the orthogonal least-squares method proposed by \citep{chen1991ols} for which efficient forward selection methods are also known. In the future, we plan to investigate how well approaches similar to our feature selection algorithm could perform on the tasks of reduced set or center selection. 

\section{Conclusion}

We propose greedy regularized least-squares, a novel training algorithm for 
sparse linear predictors. The predictors learned by the algorithm are
equivalent with those obtained by performing a greedy forward feature
selection with leave-one-out (LOO) criterion for regularized least-squares
(RLS), also known as the least-squares support vector machine or ridge
regression. That is, the algorithm works like a wrapper type of feature
selection method which starts from the empty feature set, and on each
iteration adds the feature whose addition provides the best LOO performance.
Training a predictor with greedy RLS requires $O(\desiredfcount\tsize\fsize)$
time, where $\desiredfcount$ is the number of non-zero entries in the
predictor, $\tsize$ is the number of training examples, and $\fsize$ is the
original number of features in the data. This is in contrast to the
computational complexity
$O(\min\{\desiredfcount^3\tsize^2\fsize,\desiredfcount^2\tsize^3\fsize\})$
of using the standard wrapper method with LOO selection criterion in case RLS
is used as a black-box method, and the complexity
$O(\desiredfcount\tsize^2\fsize)$ of the method proposed by
\citet{ojeda2008lssvmselection}, which is the most efficient of the
previously proposed speed-ups. Hereby, greedy RLS is computationally more
efficient than the previously known feature selection methods for RLS.

We demonstrate experimentally the computational efficiency of greedy RLS
compared to the best previously proposed implementation. In addition, we
explore the quality of the feature selection process and measure
the degree to which the LOO criterion overfits with different sizes of data
sets.

We have made freely available a software package called RLScore out of our
previously proposed RLS based machine learning
algorithms\footnote{Available at http://www.tucs.fi/RLScore}. An
implementation of the greedy RLS algorithm will also be made available as a
part of this software.

\bibliographystyle{IEEEtran}
\bibliography{MyBibliography}

\begin{thebibliography}{10}
\providecommand{\url}[1]{#1}
\csname url@samestyle\endcsname
\providecommand{\newblock}{\relax}
\providecommand{\bibinfo}[2]{#2}
\providecommand{\BIBentrySTDinterwordspacing}{\spaceskip=0pt\relax}
\providecommand{\BIBentryALTinterwordstretchfactor}{4}
\providecommand{\BIBentryALTinterwordspacing}{\spaceskip=\fontdimen2\font plus
\BIBentryALTinterwordstretchfactor\fontdimen3\font minus
  \fontdimen4\font\relax}
\providecommand{\BIBforeignlanguage}[2]{{%
\expandafter\ifx\csname l@#1\endcsname\relax
\typeout{** WARNING: IEEEtran.bst: No hyphenation pattern has been}%
\typeout{** loaded for the language `#1'. Using the pattern for}%
\typeout{** the default language instead.}%
\else
\language=\csname l@#1\endcsname
\fi
#2}}
\providecommand{\BIBdecl}{\relax}
\BIBdecl

\bibitem{guyon2003introduction}
I.~Guyon and A.~Elisseeff, ``An introduction to variable and feature
  selection,'' \emph{Journal of Machine Learning Research}, vol.~3, pp.
  1157--1182, 2003.

\bibitem{kohavi1997wrappers}
R.~Kohavi and G.~H. John, ``Wrappers for feature subset selection,''
  \emph{Artificial Intelligence}, vol.~97, no. 1-2, pp. 273--324, 1997.

\bibitem{keerthi2007fasttrack}
S.~S. Keerthi and S.~K. Shevade, ``A fast tracking algorithm for generalized
  lars/lasso,'' \emph{IEEE Transactions on Neural Networks}, vol.~18, no.~6,
  pp. 1826--1830, 2007.

\bibitem{john94irrelevant}
G.~H. John, R.~Kohavi, and K.~Pfleger, ``Irrelevant features and the subset
  selection problem,'' in \emph{Proceedings of the Eleventh International
  Conference on Machine Learning}, W.~W. Cohen and H.~Hirsch, Eds.\hskip 1em
  plus 0.5em minus 0.4em\relax San Fransisco, CA: Morgan Kaufmann Publishers,
  1994, pp. 121--129.

\bibitem{hoerl1970ridge}
A.~E. Hoerl and R.~W. Kennard, ``Ridge regression: Biased estimation for
  nonorthogonal problems,'' \emph{Technometrics}, vol.~12, pp. 55--67, 1970.

\bibitem{poggio1990networks}
T.~Poggio and F.~Girosi, ``Networks for approximation and learning,''
  \emph{Proceedings of the {IEEE}}, vol.~78, no.~9, 1990.

\bibitem{saunders1998rrdual}
C.~Saunders, A.~Gammerman, and V.~Vovk, ``Ridge regression learning algorithm
  in dual variables,'' in \emph{Proceedings of the Fifteenth International
  Conference on Machine Learning}.\hskip 1em plus 0.5em minus 0.4em\relax San
  Francisco, CA, USA: Morgan Kaufmann Publishers Inc., 1998, pp. 515--521.

\bibitem{suykens1999lssvm}
J.~A.~K. Suykens and J.~Vandewalle, ``Least squares support vector machine
  classifiers,'' \emph{Neural Processing Letters}, vol.~9, no.~3, pp. 293--300,
  1999.

\bibitem{suykens2002lssvmbook}
J.~Suykens, T.~{Van Gestel}, J.~{De Brabanter}, B.~{De Moor}, and
  J.~Vandewalle, \emph{Least Squares Support Vector Machines}.\hskip 1em plus
  0.5em minus 0.4em\relax World Scientific Pub. Co., Singapore, 2002.

\bibitem{rifkin2003rls}
R.~Rifkin, G.~Yeo, and T.~Poggio, ``Regularized least-squares classification,''
  in \emph{Advances in Learning Theory: Methods, Model and Applications}, ser.
  NATO Science Series III: Computer and System Sciences, J.~Suykens,
  G.~Horvath, S.~Basu, C.~Micchelli, and J.~Vandewalle, Eds.\hskip 1em plus
  0.5em minus 0.4em\relax Amsterdam: IOS Press, 2003, vol. 190, ch.~7, pp.
  131--154.

\bibitem{poggio2003mathematics}
T.~Poggio and S.~Smale, ``The mathematics of learning: Dealing with data,''
  \emph{Notices of the American Mathematical Society (AMS)}, vol.~50, no.~5,
  pp. 537--544, 2003.

\bibitem{vapnik1979estimation}
V.~Vapnik, \emph{Estimation of Dependences Based on Empirical Data [in
  Russian]}.\hskip 1em plus 0.5em minus 0.4em\relax Moscow, Soviet Union:
  Nauka, 1979, (English translation: Springer, New York, 1982).

\bibitem{wahba1990spline}
G.~Wahba, \emph{Spline Models for Observational Data}.\hskip 1em plus 0.5em
  minus 0.4em\relax Philadelphia, USA: Series in Applied Mathematics, Vol.~59,
  SIAM, 1990.

\bibitem{green1994nonparametric}
P.~Green and B.~Silverman, \emph{Nonparametric Regression and Generalized
  Linear Models, {A} Roughness Penalty Approach}.\hskip 1em plus 0.5em minus
  0.4em\relax London, UK: Chapman and Hall, 1994.

\bibitem{shao1993linearcv}
J.~Shao, ``Linear model selection by cross-validation,'' \emph{Journal of the
  American Statistical Association}, vol.~88, no. 422, pp. 486--494, 1993.

\bibitem{zhang1993mcv}
P.~Zhang, ``Model selection via multifold cross validation,'' \emph{The Annals
  of Statistics}, vol.~21, no.~1, pp. 299--313, 1993.

\bibitem{tang2006gene}
E.~K. Tang, P.~N. Suganthan, and X.~Yao, ``Gene selection algorithms for
  microarray data based on least squares support vector machine,'' \emph{BMC
  Bioinformatics}, vol.~7, p.~95, 2006.

\bibitem{ying2006fastloo}
Z.~Ying and K.~C. Keong, ``Fast leave-one-out evaluation for dynamic gene
  selection,'' \emph{Soft Computing}, vol.~10, no.~4, pp. 346--350, 2006.

\bibitem{ojeda2008lssvmselection}
F.~Ojeda, J.~A. Suykens, and B.~D. Moor, ``Low rank updated {LS-SVM}
  classifiers for fast variable selection,'' \emph{Neural Networks}, vol.~21,
  no. 2-3, pp. 437 -- 449, 2008, advances in Neural Networks Research: IJCNN
  '07.

\bibitem{searle1982matrixalgebra}
S.~R. Searle, \emph{Matrix Algebra Useful for Statistics}.\hskip 1em plus 0.5em
  minus 0.4em\relax New York, NY: John Wiley and Sons, Inc., 1982.

\bibitem{muller2001anintroduction}
K.-R. M{\"u}ller, S.~Mika, G.~R{\"a}tsch, K.~Tsuda, and B.~Sch{\"o}lkopf, ``An
  introduction to kernel-based learning algorithms,'' \emph{IEEE Transactions
  on Neural Networks}, vol.~12, pp. 181--201, 2001.

\bibitem{scholkopf2002kernels}
B.~Sch{\"o}lkopf and A.~J. Smola, \emph{Learning with kernels}.\hskip 1em plus
  0.5em minus 0.4em\relax MIT Press, Cambridge, MA, 2002.

\bibitem{shawetaylor2004kernel}
J.~Shawe-Taylor and N.~Cristianini, \emph{Kernel Methods for Pattern
  Analysis}.\hskip 1em plus 0.5em minus 0.4em\relax Cambridge: Cambridge
  University Press, 2004.

\bibitem{rifkin2007notes}
R.~Rifkin and R.~Lippert, ``Notes on regularized least squares,'' Massachusetts
  Institute of Technology, Tech. Rep. MIT-CSAIL-TR-2007-025, 2007.

\bibitem{russell1995ai}
S.~Russell and P.~Norvig, \emph{Artificial Intelligence: A Modern
  Approach}.\hskip 1em plus 0.5em minus 0.4em\relax Prentice Hall, 1995.

\bibitem{bottou2007svmsolvers}
L.~Bottou and C.-J. Lin, ``Support vector machine solvers,'' in
  \emph{Large-Scale Kernel Machines}, ser. Neural Information Processing,
  L.~Bottou, O.~Chapelle, D.~DeCoste, and J.~Weston, Eds.\hskip 1em plus 0.5em
  minus 0.4em\relax Cambridge, MA, USA: MIT Press, 2007, pp. 1--28.

\bibitem{pahikkala06rls}
T.~Pahikkala, J.~Boberg, and T.~Salakoski, ``Fast n-fold cross-validation for
  regularized least-squares,'' in \emph{Proceedings of the Ninth Scandinavian
  Conference on Artificial Intelligence (SCAI 2006)}, T.~Honkela, T.~Raiko,
  J.~Kortela, and H.~Valpola, Eds.\hskip 1em plus 0.5em minus 0.4em\relax
  Espoo, Finland: Otamedia, 2006, pp. 83--90.

\bibitem{an2007lssvmcv}
S.~An, W.~Liu, and S.~Venkatesh, ``Fast cross-validation algorithms for least
  squares support vector machine and kernel ridge regression,'' \emph{Pattern
  Recognition}, vol.~40, no.~8, pp. 2154--2162, 2007.

\bibitem{pudil1994floating}
P.~Pudil, J.~Novovi\v{c}ov\'{a}, and J.~Kittler, ``Floating search methods in
  feature selection,'' \emph{Pattern Recogn. Lett.}, vol.~15, no.~11, pp.
  1119--1125, 1994.

\bibitem{li2006piecewise}
J.~Li, M.~T. Manry, P.~L. Narasimha, and C.~Yu, ``Feature selection using a
  piecewise linear network,'' \emph{IEEE Transactions on Neural Networks},
  vol.~17, no.~5, pp. 1101--1115, 2006.

\bibitem{zhang2009adaptive}
T.~Zhang, ``Adaptive forward-backward greedy algorithm for sparse learning with
  linear models,'' in \emph{Advances in Neural Information Processing Systems
  21}, D.~Koller, D.~Schuurmans, Y.~Bengio, and L.~Bottou, Eds., 2009, pp.
  1921--1928.

\bibitem{pahikkala2007rankrls}
T.~Pahikkala, E.~Tsivtsivadze, A.~Airola, J.~Boberg, and T.~Salakoski,
  ``Learning to rank with pairwise regularized least-squares,'' in \emph{SIGIR
  2007 Workshop on Learning to Rank for Information Retrieval}, T.~Joachims,
  H.~Li, T.-Y. Liu, and C.~Zhai, Eds., 2007, pp. 27--33.

\bibitem{pahikkala2009ranking}
T.~Pahikkala, E.~Tsivtsivadze, A.~Airola, J.~Boberg, and J.~J{\"a}rvinen, ``An
  efficient algorithm for learning to rank from preference graphs,''
  \emph{Machine Learning}, vol.~75, no.~1, pp. 129--165, 2009.

\bibitem{smola00sparse}
A.~J. Smola and B.~Sch{\"o}lkopf, ``Sparse greedy matrix approximation for
  machine learning,'' in \emph{Proceedings of the Seventeenth International
  Conference on Machine Learning (ICML 2000)}, P.~Langley, Ed.\hskip 1em plus
  0.5em minus 0.4em\relax San Francisco, CA: Morgan Kaufmann Publishers Inc.,
  2000, pp. 911--918.

\bibitem{jiao2007fastsparse}
L.~Jiao, L.~Bo, and L.~Wang, ``Fast sparse approximation for least squares
  support vector machine,'' \emph{IEEE Transactions on Neural Networks},
  vol.~18, no.~3, pp. 685--697, 2007.

\bibitem{chen1991ols}
S.~Chen, C.~F.~N. Cowan, and P.~M. Grant, ``Orthogonal least squares learning
  algorithm for radial basis function networks,'' \emph{IEEE Transactions on
  Neural Networks}, vol.~2, no.~2, 1991.

\end{thebibliography}

\end{document}